\documentclass[10pt,twocolumn,letterpaper]{article}

\usepackage{iccv}
\usepackage{times}
\usepackage{epsfig}
\usepackage{graphicx}
\usepackage{amsmath}
\usepackage{amssymb}
\usepackage{booktabs}
\usepackage{pifont}
\usepackage[dvipsnames]{xcolor}
\usepackage{bbm}
\usepackage{comment}
\usepackage{wrapfig}
\usepackage{multirow}
\usepackage{mathtools}
\usepackage{adjustbox}
\usepackage[accsupp]{axessibility} 

\usepackage[pagebackref=true,breaklinks=true,letterpaper=true,colorlinks,bookmarks=false]{hyperref}

\hypersetup{
    colorlinks=true,
    filecolor=magenta,      
    urlcolor=cyan,
}

\newcommand\myfootnote[1]{%
  \begin{NoHyper}
  \renewcommand\thefootnote{}\footnote{#1}%
  \addtocounter{footnote}{-1}%
  \end{NoHyper}
}

\iccvfinalcopy 

\def\iccvPaperID{7201} 

\ificcvfinal\fi

\newcommand{\slfrac}[2]{\left.#1\middle/#2\right.}

\begin{document}

\title{Blending-NeRF: Text-Driven Localized Editing in Neural Radiance Fields}

\author{Hyeonseop Song$^1$\textsuperscript{*}
\and
Seokhun Choi$^1$\textsuperscript{*}
\and
Hoseok Do$^1$
\and
Chul Lee$^1$
\and
\stepcounter{footnote}{Taehyeong Kim}$^2$\textsuperscript{\dag}
\and
{\textsuperscript{1}AI Lab, CTO Division, LG Electronics, Republic of Korea}\\
{\textsuperscript{2}Dept. of Biosystems Engineering, Seoul National University, Republic of Korea}\\
{\small{\{hyeonseop.song, seokhun.choi, hoseok.do, clee.lee\}@lge.com}
\qquad\small{taehyeong.kim@snu.ac.kr}
}
}
\maketitle
\ificcvfinal\thispagestyle{empty}\fi
\myfootnote{\textsuperscript{*}Equal contribution.}
\myfootnote{\textsuperscript{\dag}Corresponding author. Partially conducted at LG Electronics.}

\begin{abstract}
Text-driven localized editing of 3D objects is particularly difficult as locally mixing the original 3D object with the intended new object and style effects without distorting the object’s form is not a straightforward process. To address this issue, we propose a novel NeRF-based model, Blending-NeRF, which consists of two NeRF networks: pretrained NeRF and editable NeRF. Additionally, we introduce new blending operations that allow Blending-NeRF to properly edit target regions which are localized by text. By using a pretrained vision-language aligned model, CLIP, we guide Blending-NeRF to add new objects with varying colors and densities, modify textures, and remove parts of the original object. Our extensive experiments demonstrate that Blending-NeRF produces naturally and locally edited 3D objects from various text prompts. Our project page is available at \href{http://seokhunchoi.github.io/Blending-NeRF}{https://seokhunchoi.github.io/Blending-NeRF}
\end{abstract}

\section{Introduction}
\label{sec:intro}
3D image synthesis and related technologies are greatly impacting industries such as art, product design, and animation. While recent 3D image synthesis techniques like Neural Radiance Field (NeRF)~\cite{mildenhall2021nerf} have opened up new applications for 3D content production~\cite{jain2022zero, lee2022understanding, poole2022dreamfusion} at scale, their ability to enable precise and localized editing of object shapes and colors remains a challenge for broader adoption. Often time, a more localized and granular editing of 3D objects, especially attaching or removing certain objects of certain styles, is still difficult and costly in spite of several recent attempts at 3D object editing~\cite{chen2022tango, liu2021editing, michel2022text2mesh, wang2022clip, yuan2022nerf}. 

\begin{figure}[t]
\centering
\begin{minipage}[b]{0.45\linewidth}
  \centering
  \centerline{\includegraphics[width=\linewidth]{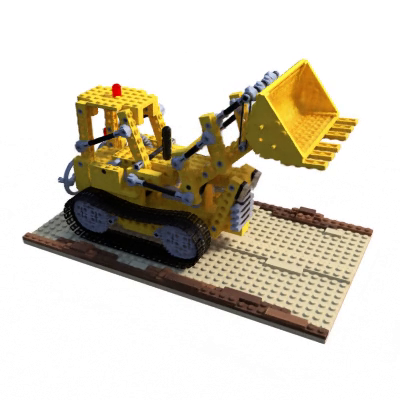}}
  \vskip -0.23in
  \centerline{\small{(a) \textit{bulldozer} (original)}}\medskip
\end{minipage}
\begin{minipage}[b]{.45\linewidth}
  \centering
  \centerline{\includegraphics[width=\linewidth]{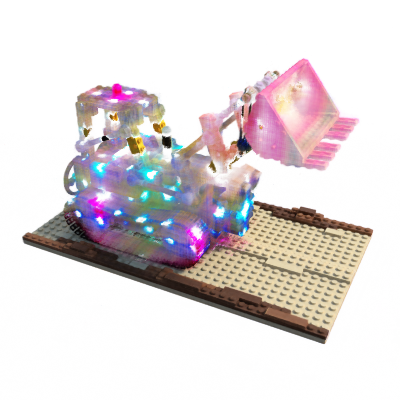}}
  \vskip -0.23in
  \centerline{\small{(b) \textit{twinkle bulldozer}}}\medskip
\end{minipage}
\vfill
\vskip -0.050in
\begin{minipage}[b]{.45\linewidth}
  \centering
  \centerline{\includegraphics[width=\linewidth]{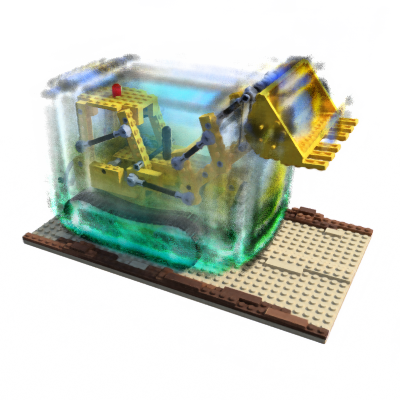}}
  \vskip -0.23in
  \centerline{\small{(c) \textit{bulldozer in glass}}}\medskip
\end{minipage}
\begin{minipage}[b]{.45\linewidth}
  \centering
  \centerline{\includegraphics[width=\linewidth]{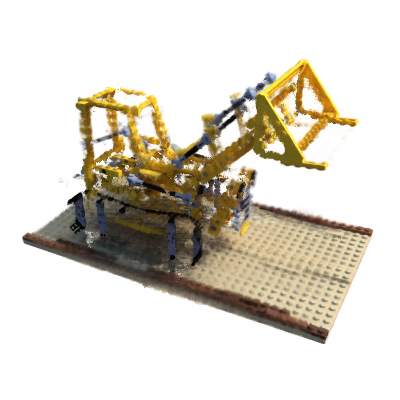}}
  \vskip -0.23in
  \centerline{\small{(d) \textit{bulldozer frame}}}\medskip
\end{minipage}

\caption{Representative results of text-driven localized object editing using our method. 
(a) Bulldozer is the original object, and each editing is performed by (b) color change, (c) density addition, and (d) density removal operations.
}
\label{fig:main_examples}
\vspace{-7px}
\end{figure}

Previous attempts, such as EditNeRF~\cite{liu2021editing} and NeRF-Editing~\cite{yuan2022nerf}, only offer limited and non-versatile editing options, while Text2Mesh~\cite{michel2022text2mesh} and TANGO~\cite{chen2022tango} allow only simple texture and shallow shape transformations of entire 3D objects. CLIP-NeRF~\cite{wang2022clip} propose a generative method with disentangled conditional NeRF for object editing but it requires a large volume of training data for the targeted editing category and is hard to edit only the desired part of objects locally. They present an additional approach, fine-tuning a single NeRF per scene with a CLIP-driven objective, which can edit object appearance but not shape well.

To achieve effective and practical localized editing of 3D objects by any text prompts at scale, it is necessary to apply style changes to specific portions of the object, including selectively changing color and locally adding and removing densities, as shown in Figure~\ref{fig:main_examples}. In this study, we propose a novel method for localized object editing that allows modification of 3D objects by text prompts, enabling full stylization including density-based localized editing. We believe that relying on the simple fine-tuning of a single NeRF to generate new densities in the low initial density area or to alter existing densities through a CLIP-driven objective is inadequate for achieving complete stylization of shapes and colors. Instead, our approach involves parameterizing specific regions in the implicit 3D volumetric representations and blending the original 3D object representation with an \textit{editable NeRF} architecture specifically trained to render the blended image naturally. We use a pretrained vision-language method like CLIPSeg \cite{lueddecke22_cvpr} to specify the area to be modified in the text input workflow.

The proposed method is based on a novel layered NeRF architecture, called Blending-NeRF, which includes a \textit{pretrained NeRF} and an \textit{editable NeRF}.
There are some studies that employ multiple NeRFs and train them simultaneously to individually reconstruct the static and dynamic components of a dynamic scene \cite{gao2021dynamic, song2023nerfplayer, wu2022d, yuan2021star}. On the other hand, our approach introduces an additional NeRF to facilitate text-based modifications in specific regions of a pretrained static scene. These modifications encompass various editing operations, including color changes, density addition, and density removal. By blending density and color from the two NeRFs, we can achieve fine-grained localized editing of 3D objects. In summary, our contributions include: 

\begin{itemize}
\vspace{-6px}
\item We propose the novel Blending-NeRF architecture that combines a \textit{pretrained NeRF} with an \textit{editable NeRF} using various objectives and training techniques. This approach allows to naturally edit the specific regions of 3D objects while preserving their original appearance.
\vspace{-16px}
\item We introduce new blending operations that capture the degree of density addition, density removal, and color alteration. Thanks to these blending operations, our method allows for precisely targeting the specific regions for localized editing and constraining the degree of object editing.
\vspace{-5px}
\item We conduct several experiments involving text-guided 3D object editing, such as editing of shape and color, and compare our approach to previous attempts and their simple extensions, showing that Blending-NeRF is both qualitatively and quantitatively superior.
\end{itemize}

\begin{figure*}[ht]
\begin{center}
\centerline{\includegraphics[width=0.99\textwidth]
{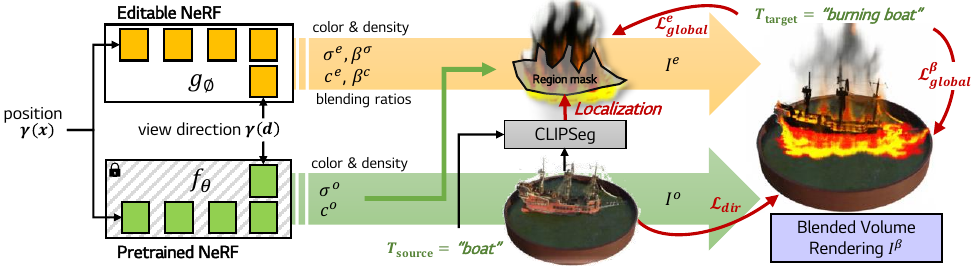}}
\vskip 0.1in
\caption{Overall architecture and main objectives. The target editing region is specified by the source text $T_{\text{source}}$ in the original rendered image $I^{o}$, and the editable NeRF $g$ is trained to render a blended image $I^{\beta}$ that matches the target text $T_{\text{target}}$. The CLIP encoders and other localized editing objectives are omitted for simplicity.
}
\vspace{-20px}
\label{fig:overview}
\end{center}
\end{figure*}

\section{Related Work}
\label{sec:related}

\paragraph{Text-Guided 3D Object Generation}
This task aims to create 3D objects from natural language descriptions. 
Recent advancements in joint embedding of images and text~\cite{radford2021learning, zhai2022lit}, text-to-image generation~\cite{ramesh2021zero, saharia2022photorealistic}, and neural rendering~\cite{barron2021mip, mildenhall2021nerf, muller2022instant, sitzmann2019deepvoxels, sun2022direct} have made it possible to generate 3D objects without 3D supervision, using only textual guidance. 
CLIP-Forge~\cite{sanghi2022clip} uses an auto-encoder and a contrastive language-image pretraining (CLIP)~\cite{radford2021learning} embedding to generate multiple object geometries for a given text query without paired text and 3D data. 
It is not that efficient, though, requiring a large unlabeled 3D dataset to train its autoencoder and to learn a latent space for shapes.
DreamField~\cite{jain2022zero} optimizes NeRF from multiple camera views to produce high-quality objects so that the CLIP embeddings of the rendered image and target text are similar. 
It improves the fidelity and visual quality of generated objects using simple geometric priors. DreamFusion~\cite{poole2022dreamfusion} uses a pretrained 2D text-to-image diffusion model~\cite{saharia2022photorealistic} and NeRF to perform text-to-3D synthesis, while CLIP-Mesh~\cite{khalid2022clip} optimizes texture, normal, and vertices position of the mesh using a differentiable renderer and CLIP. Our work utilizes NeRF and CLIP to generate 3D objects but focuses on localized editing of objects based on textual guidance, which is different from previous studies.

\vspace{-10px}
\paragraph{3D Object Editing}
Preserving the original object structure while meeting user intent is crucial in object editing tasks. Liu~\textit{et al.}~\cite{liu2021editing} proposed a conditional radiance field that enables color and shape editing by learning disentangled volumetric representation and propagating sparse 2D user scribbles over the 3D region. NeRF-Editing~\cite{yuan2022nerf} establishes the correspondence between explicit mesh representation and implicit volume representation, allowing for controllable shape deformation such as increasing or decreasing the size of 3D objects. Our method also aims at localized editing but differs in using texts as input and focusing on reshaping and restyling rather than simple modifications~\cite{liu2021editing} or geometric transformations~\cite{yuan2022nerf}.

Text-driven 3D object editing methods have also been studied. Text2Mesh~\cite{michel2022text2mesh} and TANGO~\cite{chen2022tango} edit the style of 3D objects with the supervision of CLIP. Text2Mesh stylizes a 3D mesh by predicting color and local geometries for a given target text prompt. TANGO enables photorealistic 3D style transfer by automatically predicting reflectance effects according to a text prompt without task-specific training. CLIP-NeRF~\cite{wang2022clip} allows for control of global structure and appearance individually by leveraging disentangled latent representations from conditional generative models, but it requires a significant amount of 3D dataset (e.g., 150k chair images~\cite{photoshape2018} that include sofas and wood chairs for training). While CLIP-NeRF also presents a single NeRF-based editing method per scene that can edit an object's color, it has a limitation in that it cannot edit its density well. That is, it fails to achieve satisfying results while editing the shape of a single NeRF by a text prompt. Our approach overcomes such limitations, allowing full stylization to the specific regions as demonstrated in our experiments.

\section{Background}
\subsection{Neural Radiance Field}
\label{sec:nerf} 
Neural Radiance Field (NeRF) \cite{mildenhall2021nerf} implicitly represents a 3D scene with a multi-layer perceptron (MLP) which produces a density and a color for a queried ray point sample. Specifically, given a camera ray $\mathbf{r}(t)=\mathbf{o}+t\mathbf{d}$ passing through a image pixel, depth $t \in [t_{\text{near}}, t_{\text{far}}]$, and camera center $\mathbf{o}$, it takes as input a 3D position $\mathbf{x}$ and viewing direction $\mathbf{d}$ to produce a density $\sigma^o \in [0, \infty)$ and a color $\mathbf{c}^o \in [0, 1]^3$:
\begin{equation}
\label{eq:NeRF_mlp}
(\sigma^o, \mathbf{c}^o) = f_\theta(\gamma(\mathbf{x}), \gamma(\mathbf{d})),
\end{equation}
where $\theta$ is the parameterized network weights, and $\gamma$ is a positional encoding. Then, NeRF estimates the expected color of a ray by using quadrature with $K$ sampled points:
\begin{equation}
\label{eq:original_color}
\hat{\mathbf{C}}^o(\mathbf{r})=\sum_{k=1}^{K}T_k^o\alpha_k^o\mathbf{c}_k^o,
\end{equation}
where $T_k^o=\prod_{k'=1}^{k-1}(1-\alpha_{k'}^o)$ is a transmittance \cite{10.1145/800031.808606} with an alpha value $\alpha_k^o=1-\exp(-\sigma_{k}^o\delta_{k})$ and $\delta_k=t_{k+1}-t_k$ is a distance between two adjacent sampled points on a ray. An accumulated opacity of a ray also can be estimated as:
\begin{equation}
\label{eq:NeRF_rendering_opacity}
\hat{E}^o_{\text{acc}}(\mathbf{r})=\sum_{k=1}^{K}T_k^o\alpha_k^o.
\end{equation}

\subsection{Connecting Text and Images}
There have been various studies on vision-text joint representation learning methods \cite{jia2021scaling, li2021supervision, zhai2022lit} including CLIP \cite{radford2021learning}. CLIP's image and text encoders are pretrained on large dataset to ensure that the representation vectors of image-text pairs match well. Based on these aligned representations, CLIP losses (\ie, global \cite{patashnik2021styleclip} and directional \cite{gal2022stylegan} CLIP loss) are widely used in text-guided image editing \cite{avrahami2022blended, bar2022text2live, gal2022stylegan, kim2022diffusionclip, patashnik2021styleclip, wang2022clip}. The global CLIP loss $\mathcal{L}_{\text{global}}(I,T)$ minimizes the cosine distance between an image $I$ and a text $T$ in the CLIP embedding space:
\begin{equation}
\label{eq:global_clip_loss}
\mathcal{L}_{\text{global}}(I,T) = \mathcal{D}_{\cos}(E_{\text{img}}(I),E_{\text{txt}}(T)),
\end{equation}
where $E_{\text{img}}(\cdot)$ and $E_{\text{txt}}(\cdot)$ are the image and text encoder of CLIP, and $\mathcal{D}_{\cos}$ is the cosine distance. In another way, the directional CLIP loss $\mathcal{L}_{\text{dir}}$ controls the direction of change for the image embedding vector. This method is known to prevent mode-collapsed problems \cite{gal2022stylegan}, which is defined as:
\begin{equation}
\label{eq:dir_clip_loss}
\mathcal{L}_{\text{dir}}(I_{\text{target}},T_{\text{target}},I_{\text{source}},T_{\text{source}}) = \mathcal{D}_{\cos}(\triangle I,\triangle T),
\end{equation}
where $\triangle I = E_{\text{img}}(I_{\text{target}}) - E_{\text{img}}(I_{\text{source}})$ and $\triangle T = E_{\text{txt}}(T_{\text{target}}) - E_{\text{txt}}(T_{\text{source}})$. Here, $I_{\text{target}}$ and $T_{\text{target}}$ are target edited image and its text description, and $I_{\text{source}}$ and $T_{\text{source}}$ are original image and its text description. We use both CLIP losses for text-driven object editing.

In addition, CLIP is also used for text or image-driven segmentation tasks in a zero-shot manner \cite{ding2022decoupling, li2022language, lueddecke22_cvpr, wang2022cris, xu2021simple}. We utilize CLIPSeg \cite{lueddecke22_cvpr} to get the target image region for a queried text for localized object editing.

\section{Method}
Our goal is to locally edit the pretrained NeRF model with the natural language as guidance. To this end, we propose Blending-NeRF, which consists of \textit{pretrained NeRF} $f_\theta$ for the original 3D model and \textit{editable NeRF} $g_\phi$ for object editing. The weight parameter $\theta$ is frozen, and $\phi$ is learnable. The edited scene is synthesized by blending the volumetric information of two NeRFs (Section~\ref{sec:blended_volume_rendering}).
We use two kinds of natural language prompts: \textit{source text} and \textit{target text}, describing the original and edited 3D model, respectively. Blending-NeRF performs text-driven editing using the CLIP losses with both prompts (Section~\ref{sec:text_driven_objective}). However, using only the CLIP losses is not sufficient for localized editing as it does not serve to specify the target region. Thus, during training, we specify the editing region in the original rendered scene using the source text. Simultaneously, the editable NeRF is trained to edit the target region under the guidance of localized editing objective (Section~\ref{sec:localized_editing_objective}). An overview of the proposed method is depicted in Figure \ref{fig:overview}. Note that Blending-NeRF is trained in an end-to-end manner.

\subsection{Blended Volume Rendering}
\label{sec:blended_volume_rendering}
\paragraph{Editable NeRF} 
The \textit{editable NeRF} extends NeRF to produce two blending ratios $\beta^c \in [0, 1]$ and $\beta^\sigma \in [0, 1]$ in addition to a density $\sigma^e \in [0, \infty)$ and a color $\mathbf{c}^e \in [0, 1]^3$ for seamlessly blending the ray points of two networks:
\begin{equation}
\label{eq:Blending_NeRF_mlp}
\begin{split}
(\sigma^e, \mathbf{c}^e, \beta^{\sigma}, \beta^c) &= g_\phi(\gamma(\mathbf{x}), \gamma(\mathbf{d})) \\
\sigma^{o'}&=(1-\beta^{\sigma})\sigma^{o} \\
\mathbf{c}^{o'}&=(1-\beta^c)\mathbf{c}^{o}+\beta^c \mathbf{c}^e. \\
\end{split}
\end{equation}
The density blending ratio $\beta^\sigma$ determines the amount of density $\sigma^o$ in the pretrained NeRF that is removed for object modification. Consequently, the modified original density $\sigma^{o'}$ contributes to the dominance of the editable NeRF density $\sigma^e$. Similarly, the color blending ratio $\beta^c$ controls the amount of color $\mathbf{c}^o$ modified in the pretrained NeRF. In this case, to prevent the modified original color $\mathbf{c}^{o'}$ from changing to a specific color (\eg, black or white), it is determined by mixing the editable color $\mathbf{c}^e$ by the proportion of $\beta^c$. Finally, we get $\sigma^{o'}$, $\mathbf{c}^{o'}$, $\sigma^{e}$, and $\mathbf{c}^{e}$ to blend the two NeRFs. Using these values, partial addition and removal of density, and change of color are performed on the original scene by the following blending operations. 
\vspace{-10px}
\paragraph{Blending Operations}
A previous work \cite{martin2021nerf} introduces a method for augmenting the static part with the transient part of the NeRF outputs on volume rendering to disentangle the static and transient components. Likewise, the blended color $\hat{\mathbf{C}}^{\beta}(\mathbf{r})$ of a ray can be calculated as:
\begin{equation}
\label{eq:blended_color}
\begin{split}
\hat{\mathbf{C}}^{\beta}(\mathbf{r})&=\sum_{k=1}^{K}T_k^{\beta}(\alpha_k^{o'}\mathbf{c}_k^{o'}+\alpha_k^{e}\mathbf{c}_k^e) \\
T_k^{\beta}&=\prod_{k'=1}^{k-1}(1-\alpha_{k'}^{\beta}) \\
\alpha_k^{\beta}&=1-\exp(-\sigma_{k}^{\beta}\delta_{k}), ~~ \sigma_k^{ \beta}=\sigma_k^{o'}+\sigma_k^e, \\
\end{split}
\end{equation}
where $\alpha_k^{o'}=1-\exp(-\sigma_{k}^{o'}\delta_{k})$ and $\alpha_k^e=1-\exp(-\sigma_{k}^{e}\delta_{k})$. 

In parallel, the color $\hat{\mathbf{C}}^{e}(\mathbf{r})$ of the ray for the editable NeRF is calculated as:
\begin{equation}
\label{eq:editable_color}
\hat{\mathbf{C}}^{e}(\mathbf{r})=\sum_{k=1}^{K}T_k^{\beta}(\alpha_k^{o'}\beta_k^c+\alpha_k^{e})\mathbf{c}_k^e.
\end{equation}
The colors $\hat{\mathbf{C}}^{o}$, $\hat{\mathbf{C}}^{e}$, and $\hat{\mathbf{C}}^{\beta}$ are later used to render the three images: original, editable, and blended images. 

We also define three types of accumulated opacity for the ray: $\hat{E}^{\text{add}}_{\text{acc}}$, $\hat{E}^{\text{remove}}_{\text{acc}}$, and $\hat{E}^{\text{change}}_{\text{acc}}$. The accumulated opacities are calculated as follows:
\begin{equation}
\label{eq:add_acc}
\begin{split}
\hat{E}^{\text{add}}_{\text{acc}}(\mathbf{r})&=\sum_{k=1}^{K}T_k^{\beta}\alpha_k^{e} \\
\hat{E}^{\text{remove}}_{\text{acc}}(\mathbf{r})&=\slfrac{\sum_{k=1}^{K}(T_k^{o'}-T_k^o)\alpha_k^o}{\sum_{k=1}^{K}\alpha_k^o} \\
\hat{E}^{\text{change}}_{\text{acc}}(\mathbf{r})&=\sum_{k=1}^{K}T_k^{\beta}\alpha_k^{o'}\beta_k^c,
\end{split}
\end{equation}
where $T_k^{o'}=\prod_{k'=1}^{k-1}(1-\alpha_{k'}^{o'})$. Each accumulated opacity denotes the degree of adding density, removing density, and changing color for the rendered pixel by the blending operations. Specifically, $\hat{E}^{\text{add}}_{\text{acc}}$ represents the amount of density added by the editable NeRF, and $\hat{E}^{\text{remove}}_{\text{acc}}$ represents the amount of density removed from the pretrained NeRF by the blending ratio $\beta_k^{\sigma}$. The last opacity $\hat{E}^{\text{change}}_{\text{acc}}$ means the amount of original color $\mathbf{c}_k^o$ changed by the blending operations. Note that the modifications to the object's parts that are occluded in a specific viewpoint are ignored in this operation. These accumulated opacities for the ray are used to limit the region and amount of the object editing, guided by the source text. This method, which plays an important role in localized object editing, is described in Section \ref{sec:localized_editing_objective}.

\begin{figure}[t]
\begin{center}
\centerline{\includegraphics[width=0.99\columnwidth]{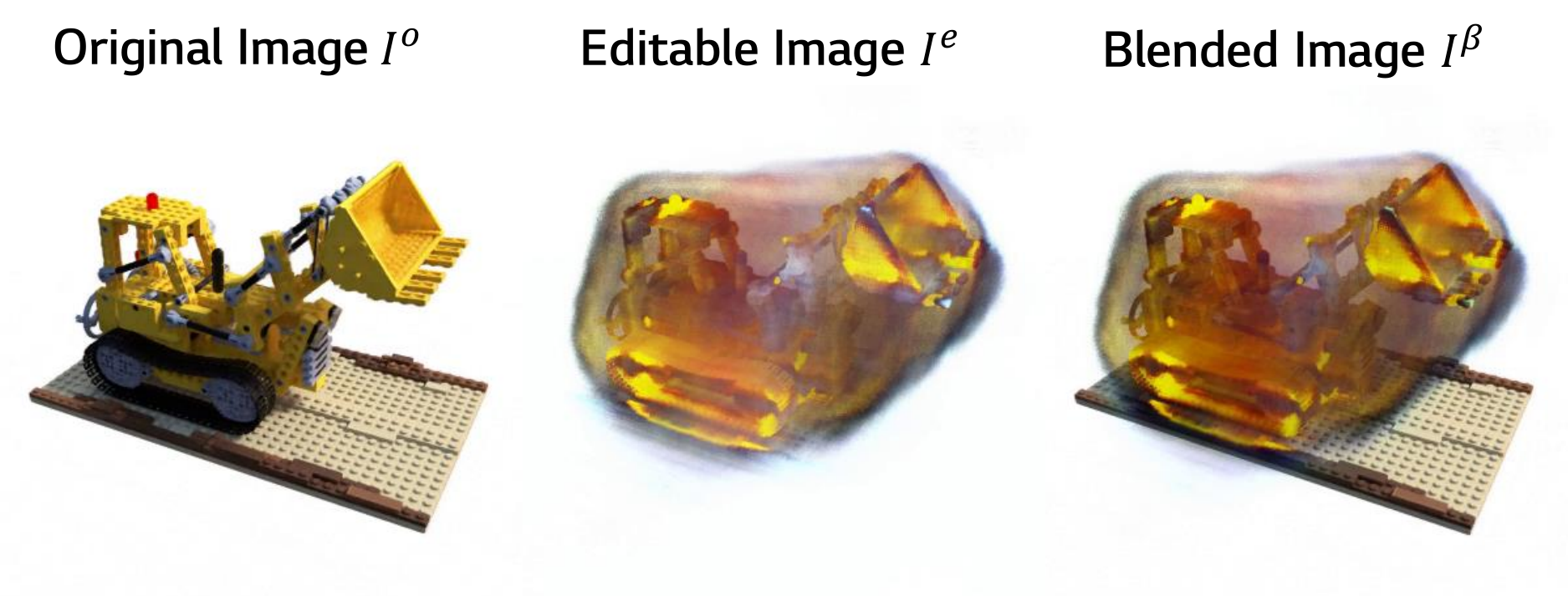}}
\caption{
The rendered images of the \textit{`bulldozer'} object edited by the target text \textit{`bulldozer amber'}.
Given a sampled camera pose, Blending-NeRF renders three types of images for training.}
\vspace{-25px}
\label{fig:rendered_image}
\end{center}
\end{figure}

\vspace{-10px}
\paragraph{Volume Rendering}
There are three types of images that are rendered with Blending-NeRF, namely original, editable, and blended images, during our localized object editing process. To train our model, we first sample a camera pose to generate these images from the sampled viewpoint. For 360$^{\circ}$ bounded scenes, we set a uniform distribution over the upper hemisphere with bounded radius \cite{jain2021putting} and sample a camera pose each training iteration. 

Given the sampled camera pose $\mathbf{p}$, the rays are also sampled at even intervals to make an image patch of size $S$ covering the entire extent of the image plane (refer to Appendix for details). Then we can obtain $S \times S$ image patches $I^o(\theta, \mathbf{p})$, $I^e(\theta, \phi, \mathbf{p})$, and $I^{\beta}(\theta, \phi, \mathbf{p})$ for original, editable, and blended images by using Eq.~(\ref{eq:original_color}), Eq.~(\ref{eq:editable_color}), and Eq.~(\ref{eq:blended_color}). Likewise, for our localized editing, the three types of opacity patches $\hat{E}^{\text{add}}_{\text{acc}}(\theta, \phi, \mathbf{p})$, $\hat{E}^{\text{remove}}_{\text{acc}}(\theta, \phi, \mathbf{p})$, and $\hat{E}^{\text{change}}_{\text{acc}}(\theta, \phi, \mathbf{p})$ of $S \times S$ size are also obtained using Eq.~(\ref{eq:add_acc}). Once trained, Blending-NeRF can render over an entire image without pixel strides at any camera pose. Examples of rendered images are shown in Figure \ref{fig:rendered_image}.

\subsection{Text-Driven Objective}
\label{sec:text_driven_objective}
We leverage the pretrained CLIP model for text-driven object editing on Blending-NeRF. For image patches $I^o(\theta,\mathbf{p})$, $I^e(\theta, \phi, \mathbf{p})$, and $I^{\beta}(\theta, \phi, \mathbf{ p})$ rendered in the previous step, we apply the global and directional CLIP losses of Eq.~(\ref{eq:global_clip_loss}) and Eq.~(\ref{eq:dir_clip_loss}): $\mathcal{L}^{e}_{\text{global}}(I^{e}(\theta, \phi, \mathbf{p}),T_{\text{target}})$, $\mathcal{L}^{\beta}_{\text{global}}(I^{\beta}(\theta, \phi, \mathbf{p}),T_{\text{target}})$ and $\mathcal{L}_{\text{dir}}(I^{\beta}(\theta, \phi, \mathbf{p}),T_{\text{target}},I^o(\theta,\mathbf{p}),T_{\text{source}})$. The global CLIP losses $\mathcal{L}_{\text{global}}^{e}$ and $\mathcal{L}^{\beta}_{\text{global}}$ make CLIP embeddings of both editable image $I^{e}$ and blended image $I^{\beta}$ close to that of target text $T_{\text{target}}$. The directional CLIP loss ensures that the direction of representation vector from the source image $I^{o}$ to the blended image $I^{\beta}$ is similar to the direction from the source text $T_{\text{source}}$ to the target text $T_{\text{target}}$. The total text-driven objective is defined as:
\begin{equation}
\label{eq:total_clip_loss}
\mathcal{L}_{\text{clip}} = \mathcal{L}_{\text{dir}} + \lambda_{\text{global}}\mathcal{L}_{\text{global}},
\end{equation}
where $\mathcal{L}_{\text{global}}=\mathcal{L}^{\beta}_{\text{global}}+\mathcal{L}_{\text{global}}^e$ is the global CLIP loss, and $\lambda_{\text{global}}$ is a hyperparameter for balancing the directional and global CLIP losses.

\vspace{-10px}
\paragraph{Image and Text Augmentation}
Before feeding image patches and text prompts to CLIP encoders, we apply image and text augmentations. Previous work~\cite{lee2022understanding} shows that applying 2D image-based augmentations can prevent adversarial generation problems when using CLIP guidance. Similarly, we augment image patches in the order of differential \cite{zhao2020diffaugment} and random perspective augmentations. We use the text templates \cite{bar2022text2live} to augment $\textit{T}_{\text{source}}$ and $\textit{T}_{\text{target}}$.

\subsection{Localized Editing Objective}
\label{sec:localized_editing_objective}
The text-driven objective can guide Blending-NeRF to edit the original object to match the meaning of a given target text $T_{\text{target}}$.
However, with the CLIP losses alone, it is challenging to specify the region and amount of editing. Thus, we employ a text-guided semantic segmentation method and the opacity patches $\hat{E}^{\text{add}}_{\text{acc}}(\theta, \phi, \mathbf{p})$, $\hat{E}^{\text{remove}}_{\text{acc}}(\theta, \phi, \mathbf{p})$, and $\hat{E}^{\text{change}}_{\text{acc}}(\theta, \phi, \mathbf{p})$. Note that precisely targeting the region for localized editing and constraining the degree of object editing is well handled by these three accumulated opacities, which capture the extent of density addition, density removal, and color alteration through our blending operations. In addition, joint optimization of localizing target region and constraining editable amount to maintain the high-fidelity results of the pre-trained NeRF is a vital factor in producing localized editing that is less prone to noise, as demonstrated throughout our experimental results in Section~\ref{sec:ablation}.

\vspace{-10px}
\paragraph{Localizing Target Region}
We use CLIPSeg~\cite{lueddecke22_cvpr} to guide the region to be edited only with a user text prompt $\textit{T}_{\text{source}}$. Specifically, we leverage zero-shot segmentation $h(I, T)$ to produce a probability map of the pixels in an image $I$ associated with the input text $T$. We first estimate region $M$, which is more likely to be $\textit{T}_{\text{source}}$ than the text \textit{`photo'} in the source image $I^o(\theta, \mathbf{p})$, as follows:
\begin{equation}
\label{eq:target_region}
M = \mathbbm{1}(h(I^o(\theta, \mathbf{p}), \textit{T}_{\text{source}}) - h(I^o(\theta, \mathbf{p}), \textit{`photo'})),
\end{equation}
where function $\mathbbm{1}(\cdot)$ pixel-wisely outputs 1 if its input is positive, and 0 otherwise. After applying $N_f$ dilation operations to $M$, we get the positive target region ${M}_{+}$ which specifies the region of interest to edit. Additionally, we specify the negative target region ${M}_{-}$ which designates the region of non-interest by applying $N_f$ dilatation operations to ${M}_{+}$ and element-wise \textit{not} operation.  Then the loss to localize the target region is:
\begin{equation}
\label{eq:loss_region}
\begin{split}
\mathcal{L}_{\text{region}}= \ & \text{MSE}([0]_{S \times S}, {M}_{-}\odot E_{\text{sum}})+\\
&\lambda_{+}\text{MSE}([1]_{S \times S}, {M}_{+}\odot E_{\text{sum}}),
\end{split}
\end{equation}
where $E_{\text{sum}}=\sum_{x=\{\text{add,remove,change}\}}\hat{E}^{x}_{\text{acc}}(\theta, \phi, \mathbf{p})$ is the pixel-wise sum of the three accumulated opacities, $\odot$ denotes pixel-wise multiplication, and $\lambda_{+}$ is a hyperparameter for balancing the two terms. The first term prevents modification outside the target region, while the second encourages editing within the target region.

\begin{figure*}[t]
\begin{center}
\centerline{\includegraphics[width=0.99\textwidth]{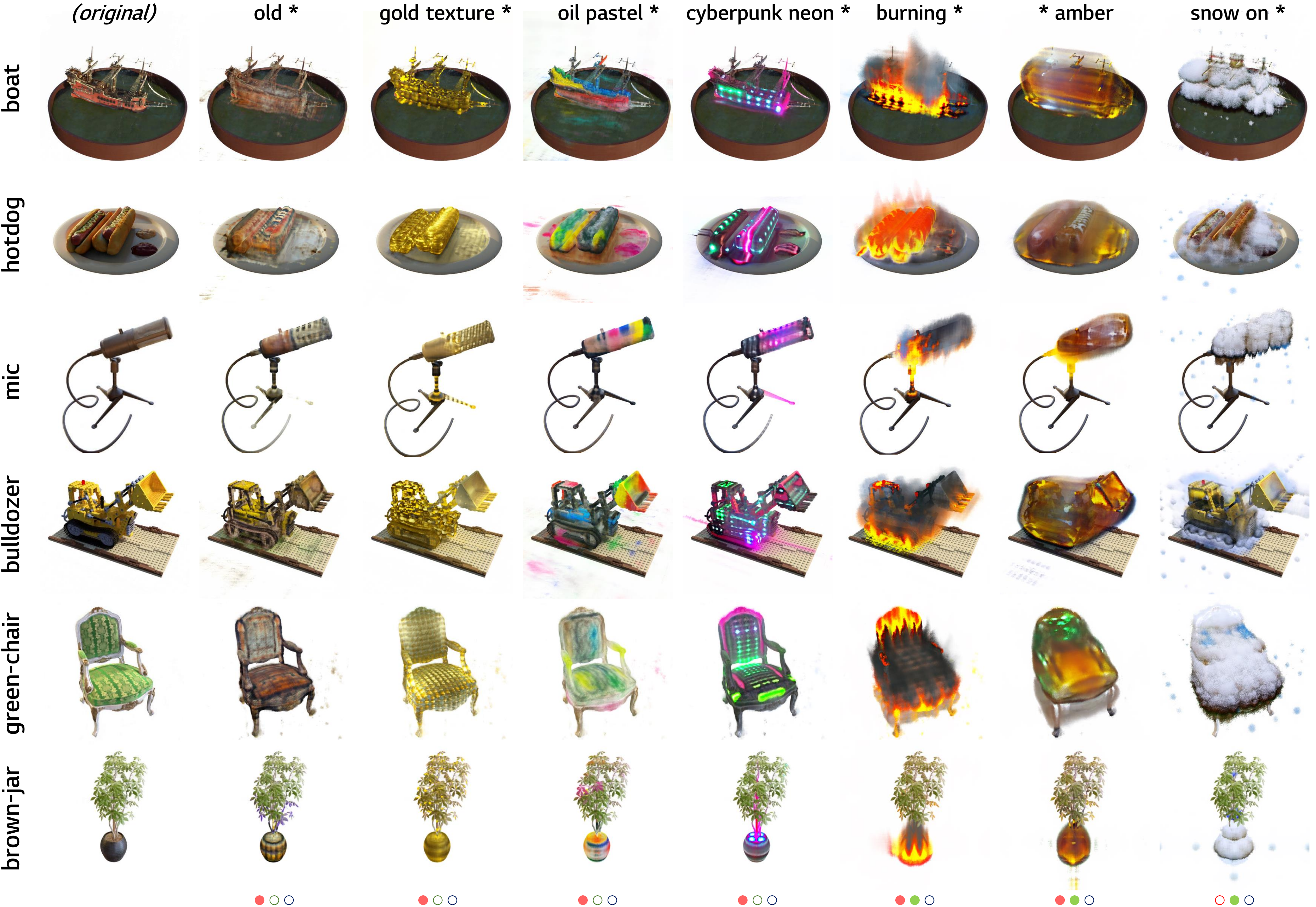}}
\caption{Examples of editing in which the common target text templates are applied to various 3D objects. The images in the first column are the original objects.
The object names are listed on the left side of the figure, and the target text templates are listed on the top of the figure. For example, the second image in the first row is an edited result with \textit{`old boat'}, which combines \textit{`boat'} and \textit{`old} *\textit{'}. We denote the {\color{RubineRed} $\bullet$} as changing colors, {\color{LimeGreen} $\bullet$} as adding densities, and {\color{NavyBlue} $\bullet$} as removing densities of the object. If two or more dots exist, the editing is performed with the corresponding case together. This notation is common to all figures.
\centerline{{\color{RubineRed} $\bullet$} : changing colors \qquad {\color{LimeGreen} $\bullet$} : adding densities \qquad {\color{NavyBlue} $\bullet$} : removing densities.}
}
\vspace{-25px}
\label{fig:main_result}
\end{center}
\end{figure*}

\begin{figure}[t]
\begin{center}
\centerline{\includegraphics[width=0.8\columnwidth]{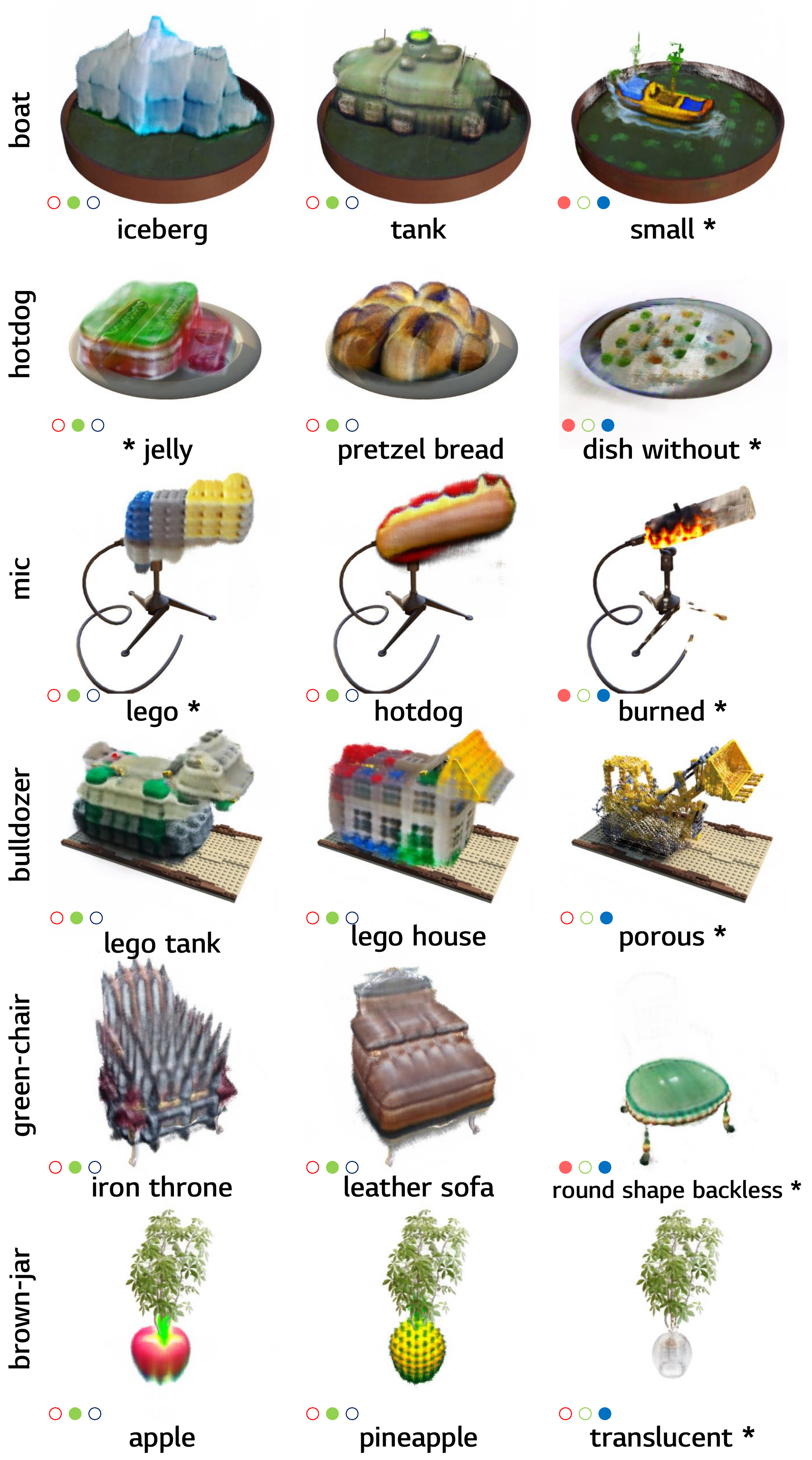}}
\caption{
Examples of editing in which the target-specific text descriptions are applied to 3D objects. The texts on the left of the images refer to the editing target on the original objects.
}
\label{fig:main_result2}
\end{center}
\vskip -25px
\end{figure}

\vspace{-10px}
\paragraph{Constraining Editable Amount}
To limit the amount of area being modified, we use an opacity loss similar to the transmittance loss in the previous work \cite{jain2022zero}. The opacity loss $\mathcal{L}_{\text{opacity}}$ is defined using the opacity patches $\hat{E}^{\text{add}}_{\text{acc}}(\theta, \phi, \mathbf{p})$, $\hat{E}^{\text{remove}}_{\text{acc}}(\theta, \phi, \mathbf{p})$, and $\hat{E}^{\text{change}}_{\text{acc}}(\theta, \phi, \mathbf{p})$ as follows:
\begin{equation}
\label{eq:loss_opacity}
\mathcal{L}_{\text{opacity}}=\sum_{x}\max(\tau^x, \text{mean}(\hat{E}^{x}_{\text{acc}}(\theta, \phi, \mathbf{p}))),
\end{equation}
where $x=\{\text{add, remove, change}\}$ and $\{\tau^x\}$ are the thresholds to limit the amount of addition and removal of density, and change of color. These thresholds are annealed for stable learning. 

We also apply the regularization loss $\mathcal{L}_{\text{reg}}$ to the opacity patch $\hat{E}^{\text{add}}_{\text{acc}}(\theta, \phi, \mathbf{p})$ to avoid adding ambiguous densities:
\begin{equation}
\label{eq:loss_reg}
\mathcal{L}_{\text{reg}}=-\text{mean}(F(\hat{E}^{\text{add}}_{\text{acc}}(\theta, \phi, \mathbf{p}))),
\end{equation}
where $F(z)=z\log_2 z + (1-z)\log_2(1-z)$ is the binary entropy function. Note that we use stop gradients to ensure that localized editing losses do not indiscriminately affect training. In particular, the losses for opacity patch $\hat{E}^{\text{add}}_{\text{acc}}(\theta, \phi, \mathbf{p})$ to add density are only concerned with the backpropagation by editable density $\sigma^{e}$. Likewise, the losses by opacity patches $\hat{E}^{\text{remove}}_{\text{acc}}(\theta, \phi, \mathbf{p})$ and $\hat{E}^{\text{change}}_{\text{acc}}(\theta, \phi, \mathbf{p})$ propagate only to $\beta^{\sigma}$ and $\beta^c$, respectively.

Finally, our total objective $\mathcal{L}_{\text{total}}$ for text-driven localized object editing is: 
\begin{equation}
\label{eq:loss_total}
\mathcal{L}_{\text{total}}=\mathcal{L}_{\text{clip}}+\lambda_1\mathcal{L}_{\text{region}}+\lambda_2\mathcal{L}_{\text{opacity}}+\lambda_3\mathcal{L}_{\text{reg}},
\end{equation}
where $\lambda_1$, $\lambda_2$, and $\lambda_3$ are hyperparameters to balance losses.

\section{Experiments and Results}

\subsection{Implementation Details}
\label{sec:implementation_details}
The pretrained NeRF consists of an 8-layer MLP of 256 hidden units with ReLU activations as in the architecture of the originally proposed NeRF \cite{mildenhall2021nerf}. For the editable NeRF, we partially modified the original NeRF using residual blocks (see Appendix for details). We followed the same procedures in the hierarchical volume sampling of NeRF as well, but we did the importance sampling based on $T^{\beta}_k \alpha^{\beta}_k$ instead of $T^{o}_k \alpha^{o}_k$. The patch size for all images and accumulated opacities is $S=72$. We used Adam Optimizer and the learning rate is linearly decayed from $5\times10^{-4}$ to $10^{-4}$ for the first 1k iteration steps and stays at $10^{-4}$ for the remaining steps. We included the regularization loss component only after the first 1k iteration steps, once the density of the newly added object has reached a certain level of form. Regarding the hyperparameter values, we employed the following: $\lambda_{\text{global}}=0.5$, $\lambda_1=1$, $\lambda_2=2$, and $\lambda_3=0.2$. We evaluated our method using a variety of target texts and six 3D objects (ship, hotdog, mic, lego, chair, and ficus) from the Realistic Synthetic 360$^\circ$ dataset \cite{mildenhall2021nerf}.

\begin{figure}[t!]
\begin{center}
\centerline{\includegraphics[width=0.99\columnwidth]{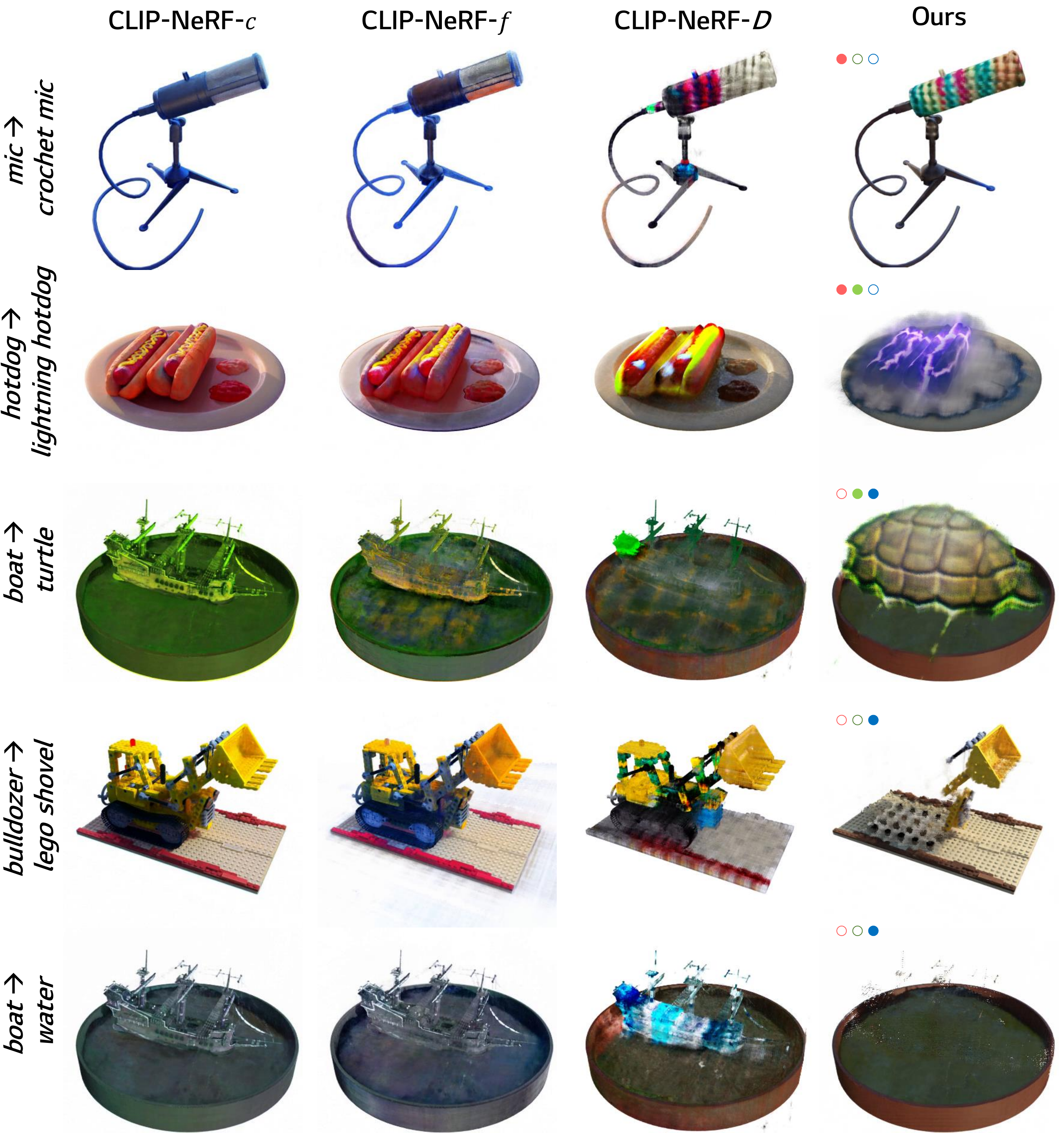}}

\vskip 0.1in
\caption{Comparison with baselines. Our method demonstrates superior ability, particularly in editing density, as evidenced by the third row where we added densities, as well as the fourth and fifth rows where we removed densities. In contrast, the competing methods fail to achieve these tasks.
}
\label{fig:qual_comparison_baseline}
\end{center}
\vskip -15px
\end{figure}

\subsection{Localized Editing}
To investigate the performance of our method for localized object editing, we performed a variety of experiments such as addition or removal of densities, and color changes to the original objects. Figure~\ref{fig:main_result} shows the edited results obtained by applying the same target text templates to all source objects. Our method clearly achieves a detailed editing of the target object while preserving the source structure. For example, when given a source text \textit{`boat'} and a target text \textit{`gold texture boat'}, a well-stylized boat with a gold texture was rendered while preserving the background. The localized editing also worked well when simultaneously adding densities to an object and changing its colors. In the examples of \textit{`burning bulldozer'} or \textit{`snow on mic'}, the appropriate ambient effect appeared naturally along with editing the target object.

We extended our experiments to a more diverse set of target texts, as shown in Figure \ref{fig:main_result2}. In particular, we performed object editing tasks to remove densities. For instance, given a source text \textit{`green-chair'} and a target text \textit{`round shape backless green-chair'}, the back and armrests of the chair were removed to achieve the editing goal.

\subsection{Comparison with Baselines}
\paragraph{Baselines}
To demonstrate the effectiveness of our approach, we compared it against three different variants of CLIP-NeRF. Wang~\textit{et al.}~\cite{wang2022clip} present a single NeRF-based editing method per scene (let’s call it CLIP-NeRF-\textit{c}). We compared our method against CLIP-NeRF-\textit{c}, which only fine-tunes its color-related layers, using officially released code. We also evaluated our method against CLIP-NeRF-\textit{f}, which fine-tunes all layers instead of just the color-related ones. Additionally, we compared our method against another variant, CLIP-NeRF-\textit{D}, which uses distilled feature fields~\cite{kobayashi2022decomposing} as a localization module using official code that fine-tunes all layers.
\begin{table}[t!]
    \begin{center}
    \begin{small}
    \resizebox{\columnwidth}{!}{%
    \footnotesize
        \begin{tabular}{c c c c c}
        \toprule
            & CLIP-NeRF-\textit{c} & CLIP-NeRF-\textit{f} & CLIP-NeRF-\textit{D} & Ours \\
        \midrule
          $D_{L_1} \downarrow$  &  \textbf{.029} & .041 & .047 & .051 \\
          $S_{\text{CLIP}} \uparrow$ & .065 & .081 & .084 & \textbf{.128} \\
          $\text{MP}_{\text{CLIP}} \uparrow$ & .063 & .077 & .080 & \textbf{.121} \\
        \bottomrule
        \end{tabular}
    }
    \end{small}
    \end{center}
\caption{
Quantitative comparison with baseline models. We measured the preservation of the original appearance ($D_{L_1}$) and the alignment with the target text ($S_{\text{CLIP}}$). We also measured the manipulative precision ($\text{MP}_{\text{CLIP}}$) to consider them both. 
}
\label{tab:quant_comp_baseline}
\end{table}

\vspace{-10px}
\paragraph{Evaluation Metric}
We evaluated the quality of text-driven object editing using the manipulative precision (MP) metric \cite{li2020manigan}. The MP metric takes into account two aspects: the preservation of the original appearance, which is measured as the L1 normalized pixel distance ($D_{L_1}$) between the original and edited image, and the alignment with the target text, which is measured by the CLIP score ($S_{\text{CLIP}}$) between the edited image and target text. The CLIP score is obtained by averaging CLIP similarity and Directional CLIP similarity~\cite{kim2022diffusionclip}. For a fair comparison, we used CLIP ViT-L/14 to calculate the CLIP score instead of CLIP ViT-B/32 used to train Blending-NeRF and the baselines. Finally, the CLIP based MP metric is defined as $\text{MP}_{\text{CLIP}} = (1 - D_{L_1}) \times S_{\text{CLIP}}$. We calculated all these metrics using 60 different scenes for each model. 
\vspace{-10px}
\paragraph{Comparisons}
We qualitatively and quantitatively compared the performance of Blending-NeRF with three variants of CLIP-NeRF. Our method outperformed all baselines, as shown in Figure~\ref{fig:qual_comparison_baseline}. The CLIP-NeRF variants were able to perform the color change task (\textit{mic} $\rightarrow$ \textit{crochet mic}), but struggled with the density change tasks. Although the localizing module helped CLIP-NeRF-\textit{D} to edit the target region well, it still had difficulty in editing the densities of the target object. As indicated by Wang~\textit{et al.}~\cite{wang2022clip}, this demonstrates that relying on the simple fine-tuning of a single NeRF to generate new densities in the low initial density area or to alter existing densities through a CLIP-driven objective is not sufficient for achieving complete localized object editing of shapes. Instead, we found in our experiments that using our novel dual NeRF architecture to blend volumetric information from two independent NeRFs, namely, pretrained NeRF capturing the original 3D model and editable NeRF capturing object editing information, and specifying the positive and negative regions to help the blended editing focus on the target regions results in more natural localized object editing. These qualitative results in Figure~\ref{fig:qual_comparison_baseline} are consistent with the superior quantitative performance of our model, measured using MP metric, as shown in Table~\ref{tab:quant_comp_baseline}.

\begin{figure}[t!]
\begin{center}
\centerline{\includegraphics[width=0.75\columnwidth]{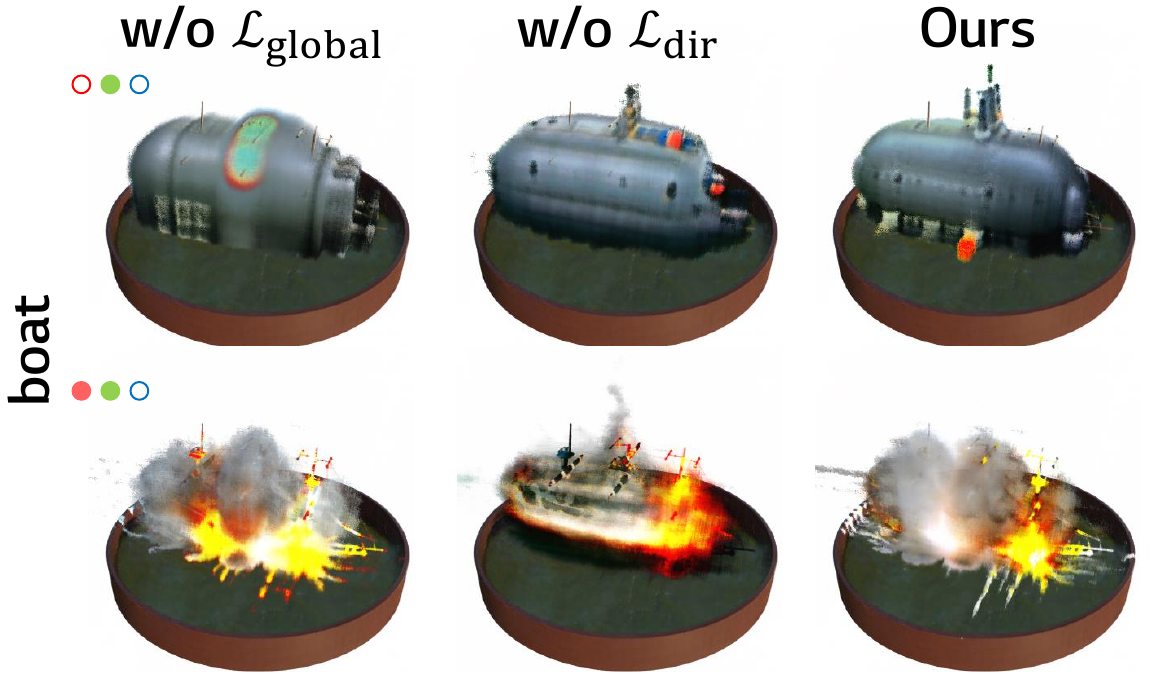}}
\vskip 0.1in
\caption{Ablation study on the text-driven objectives.
The \textit{`boat'} object is edited with \textit{`submarine'} (upper row) and \textit{`exploding boat'} (bottom row) as the target texts.}
\label{fig:ablation_clip_objective}
\end{center}
\vskip -25px
\end{figure}

\begin{figure}[t!]
\begin{center}
\centerline{\includegraphics[width=1.0\columnwidth]{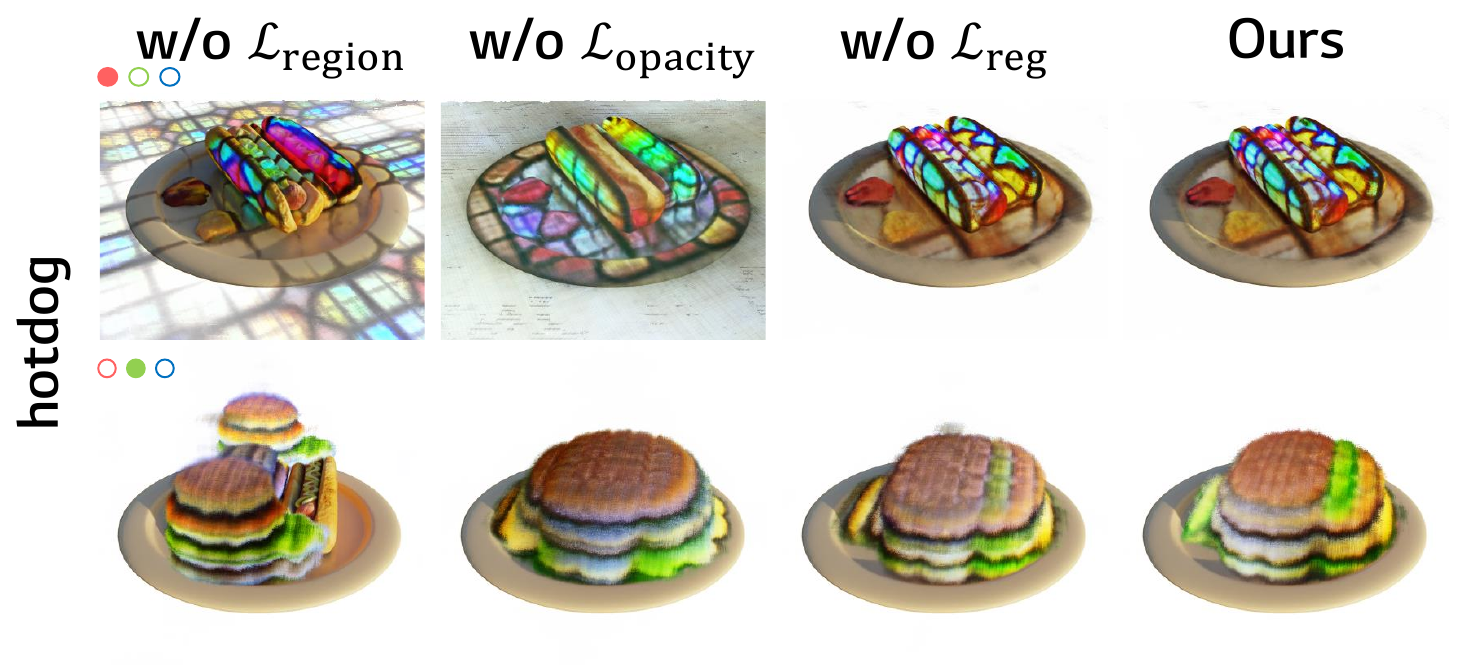}}
\caption{Ablation study on the localized editing objectives.
The \textit{`hotdog'} object is edited with \textit{`stained glass hotdog'} (upper row) and \textit{`hamburger'} (bottom row) as the target texts.}
\label{fig:ablation_localized_editing_objective}
\end{center}
\vskip -0.25in
\end{figure}

\begin{table}[t]
    \begin{center}
    \begin{small}
    \resizebox{\columnwidth}{!}{
    \footnotesize
        \begin{tabular}{c c c c c c c}
        \toprule
          &  \scriptsize{w/o} $\mathcal{L}_{\text{region}}$ & \scriptsize{w/o} $\mathcal{L}_{\text{opacity}}$ & \scriptsize{w/o} $\mathcal{L}_{\text{reg}}$ & \scriptsize{w/o} $\mathcal{L}_{\text{global}}$ & \scriptsize{w/o} $\mathcal{L}_{\text{dir}}$ & Ours \\
        \midrule
          $D_{L_1} \downarrow$  &  \textbf{.049} & .085 & .053 & .053 & .049 & .051 \\
          $S_{\text{CLIP}} \uparrow$ & .121 & .126 & .125 & .122 & .108 & \textbf{.128} \\
          $\text{MP}_{\text{CLIP}} \uparrow$ & .115 & .115 & .119 & .115 & .103 & \textbf{.121} \\
        \bottomrule
        \end{tabular}
    }
    \end{small}
    \end{center}
\caption{
Quantitative comparison with ablation models. We ablate the proposed localized editing (w/o~$\mathcal{L}_{\text{region}}$, w/o~$\mathcal{L}_{\text{opacity}}$, and w/o~$\mathcal{L}_{\text{reg}}$) and text-driven objectives (w/o~$\mathcal{L}_{\text{global}}$ and w/o~$\mathcal{L}_{\text{dir}}$).
}
\label{tab:quant_comp_ablation}
\end{table}

\subsection{Ablation Study}
\label{sec:ablation}
To validate the effect of our text-driven and localized editing losses, we compared the performance qualitatively as well as quantitatively when each loss term was removed from the total objective. We first performed an ablation study on text-driven losses, as shown in Figure \ref{fig:ablation_clip_objective}. In the \textit{`submarine'} case, when global CLIP loss was not used (w/o $\mathcal{L}_{\text{global}}$), the result was blurry with degraded quality. Similarly, in the \textit{`exploding boat'} case, when only global CLIP loss was used (w/o $\mathcal{L}_{\text{dir}}$), the explosion effect was not well expressed, resulting in a poor editing performance. These results are also consistent with the poor CLIP scores and MP metrics, as shown in Table~\ref{tab:quant_comp_ablation}. That is, using both global and directional losses enhances the overall editing quality.

We further analyzed the effect of localized editing losses on localizing the target region. As shown in Figure~\ref{fig:ablation_localized_editing_objective}, when each localized editing loss was excluded (\ie, w/o $\mathcal{L}_{\text{region}}$ or w/o $\mathcal{L}_{\text{opacity}}$), the editing regions were not well targeted or adequately constrained overall. As shown in Table~\ref{tab:quant_comp_ablation}, these results are also consistent with the low MP metric (w/o $\mathcal{L}_{\text{region}}$) and the poor preservation score (w/o $\mathcal{L}_{\text{opacity}}$). Additionally, for the w/o $\mathcal{L}_{\text{reg}}$ in the \textit{`hamburger'} case, the edited object has ambiguous boundaries. In contrast, our method edits objects with clear boundaries and less noise. This result implies that the regularization loss guides Blending-NeRF to add density distinctly, improving the MP metric as shown in Table~\ref{tab:quant_comp_ablation}. That is, our method can locally and naturally edit the target object in the original scene with only minor modifications to the regions of non-interest.

\subsection{Extendability of Blending-NeRF}
\label{sec:extendability}
\begin{figure}[t!]
\begin{center}
\centerline{\includegraphics[width=1.0\columnwidth]{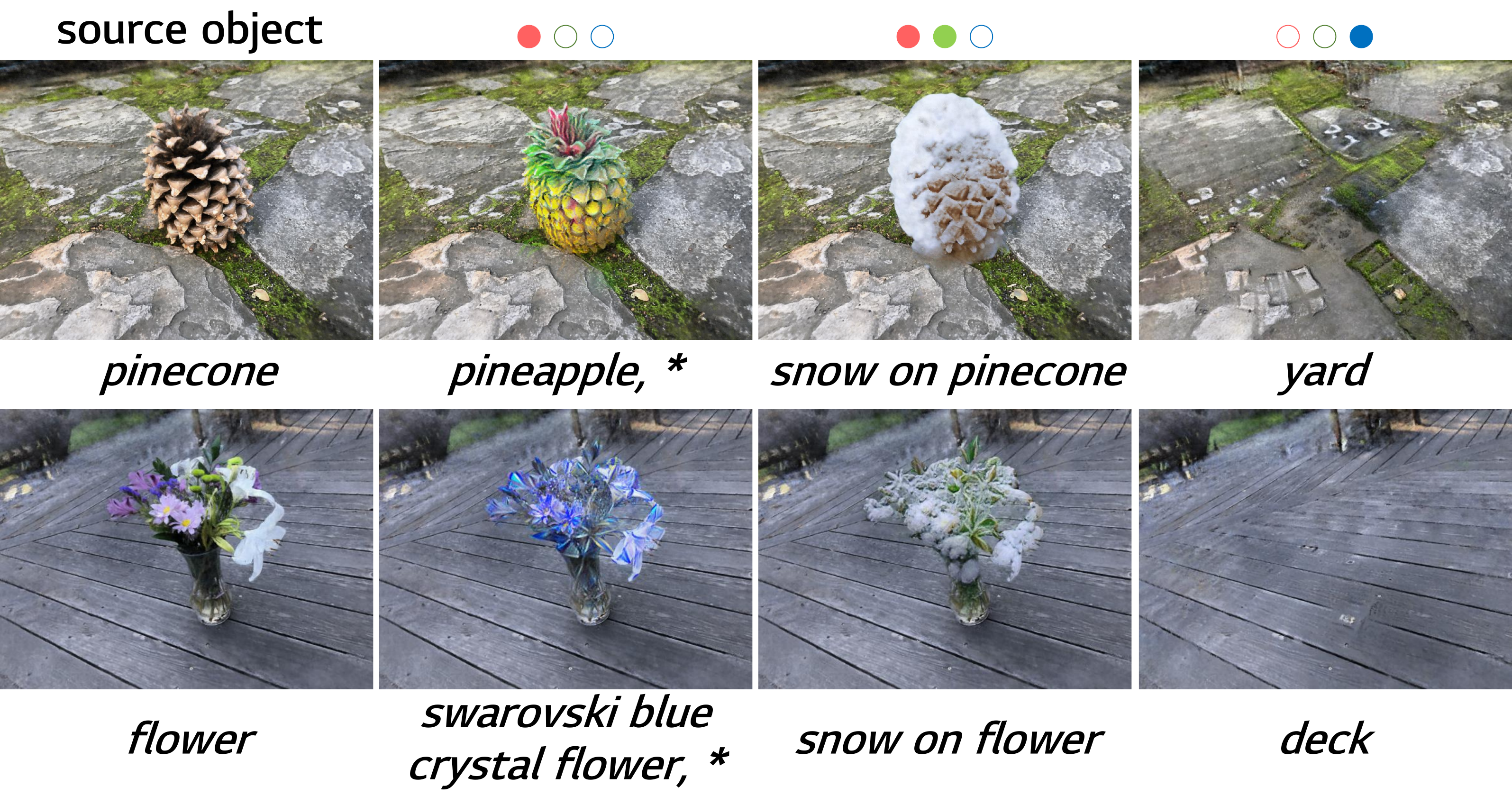}}
\caption{Examples of localized editing on the \textit{pinecone} and \textit{vasedeck} scenes. Blending-NeRF with Instant-NGP was used for these results. (* : \textit{trending on artstation})
}
\label{fig:ngp_real_scene}
\end{center}
\vskip -0.25in
\end{figure}

We investigated the extendability of the proposed method using Instant-NGP~\cite{muller2022instant} which utilizes hash grid encoding to represent a 3D scene with low computational cost. The localized editing results on real scenes~\cite{mildenhall2021nerf} in Figure \ref{fig:ngp_real_scene} demonstrate that our method can be integrated with other 3D scene representation methods such as Instant-NGP. In this experiment, Blending-NeRF was able to inherit the advantages of Instant-NGP over the originally proposed NeRF~\cite{mildenhall2021nerf} on memory efficiency and training time. For more results and implementation details, refer to Section~E of Appendix.

\subsection{Editing Operations}
\begin{figure}[h!]
\begin{center}
\centerline{\includegraphics[width=0.9\columnwidth]{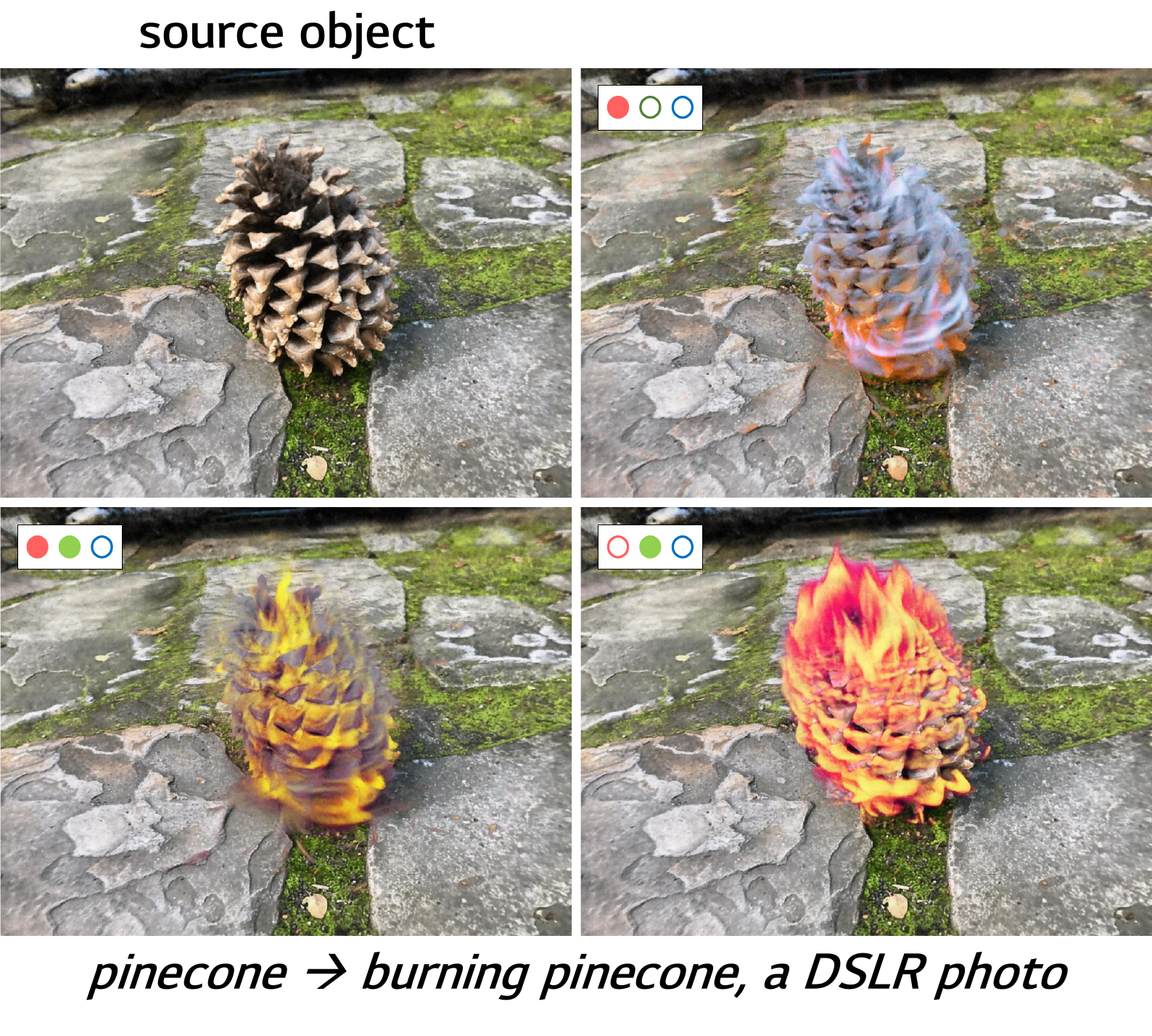}}
\caption{Experiments on using different editing operations. The source object (\textit{`pinecone'}) is edited into each object using the same target text (\textit{`burning pinecone, a DSLR photo'}) with different combinations of editing operations. Note that Blending-NeRF with Instant-NGP was used for these results.
}
\label{fig:ngp_editing_operations}
\end{center}
\vskip -25px
\end{figure}

Our approach explicitly distinguishes editing operations, such as adding and removing density, and changing color. The editing results can vary even for the same scene and text, depending on the manually specified editing operations. By choosing different combinations of the editing operations, users have the ability to achieve desired editing, as shown in Figure~\ref{fig:ngp_editing_operations}. 

\subsection{Limitations}
Our work has limitations in that the overall performance can be affected by the two off-the-shelf models, CLIPSeg and CLIP. For instance, if the segmentation of the target area by CLIPSeg is not appropriate, unedited parts may remain. This performance degradation can be mitigated by using advanced segmentation models or a potential solution described in Appendix (\ie, user-provided mask).

\begin{figure}[t!]
\begin{center}
\centerline{\includegraphics[width=1\columnwidth]{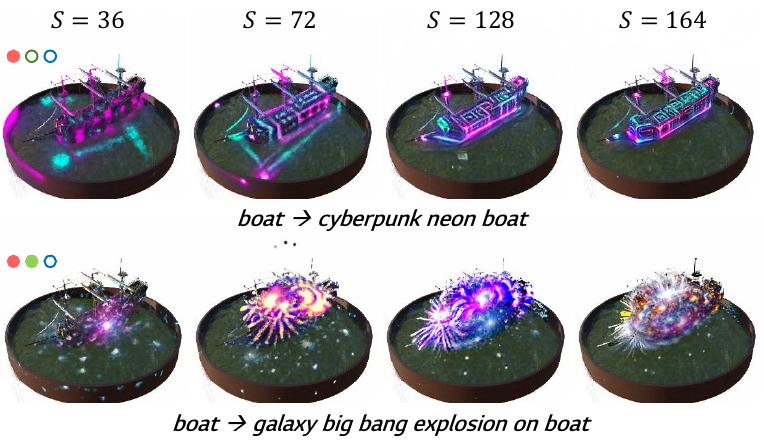}}
\vskip -0.05in
\caption{Experiments on object editing with various patch sizes. The numbers at the top denote each patch size used. The source object (\textit{`boat'}) is edited into each object using two target texts: \textit{`cyberpunk neon boat'} (top row) and \textit{`galaxy big bang explosion on boat'} (bottom row), respectively. Note that Blending-NeRF with Instant-NGP was used for these results.
}
\vskip -25px
\label{fig:ngp_patch_size}
\end{center}
\end{figure}

Additionally, we found that the limited patch size input to CLIP's image encoder can make edited results blurry. The input size of the CLIP encoder is 224, but we used a patch size of 72 due to our computational resources when we used the originally proposed NeRF~\cite{mildenhall2021nerf} as a backbone. However, this issue can be alleviated by using a memory-efficient backbone (\ie, Instant-NGP). As shown in Figure~\ref{fig:ngp_patch_size}, where we applied the proposed method to Instant-NGP as described in Section~\ref{sec:extendability}, the blurry results were improved as the patch sizes increased. Considering the trade-off between the quality improvement and the increase in computational time, we set the patch size as 128 for our experiments using Instant-NGP.

\section{Conclusion}
For text-driven localized 3D object editing, we propose Blending-NeRF, which consists of pretrained NeRF and editable NeRF. The target region for editing is specified by the source text and the original object in the pretrained NeRF. Blending-NeRF renders blended images of two NeRFs suitable for the target text by freezing the pretrained NeRF and training the editable NeRF to locally edit the original object while maintaining the overall appearance. Especially, we define three types of editing operations (\ie, adding or removing density, changing color) and use them to perform various 3D object editing. Empirical results show that our approach is superior to text-driven localized object editing. We firmly believe that the proposed method and localized object editing tasks hold practical value in neural rendering.

{\small
\bibliographystyle{ieee_fullname}
\bibliography{egbib}
}

\clearpage
\onecolumn

\clearpage
\appendix
\onecolumn

\begin{center}
\large \textbf{\\Blending-NeRF: Text-Driven Localized Editing in Neural Radiance Fields \\Supplementary Material}
\end{center}

\vskip 0.3in
\section{Implementation Details}

\subsection{Dilatation Operations}
\begin{wrapfigure}{r}{0.3\textwidth}
\begin{center}
\vskip -0.6in
\centerline{\includegraphics[width=0.3\columnwidth]{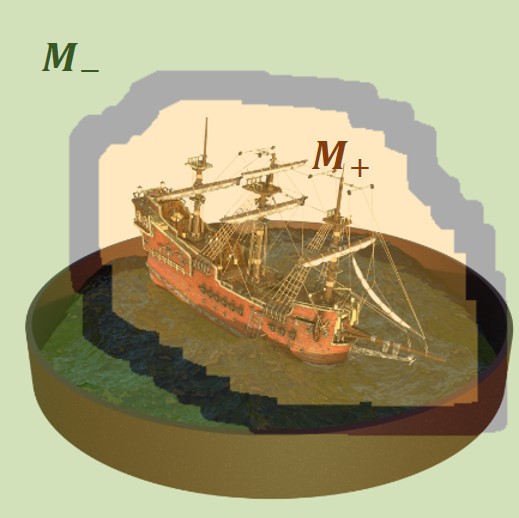}}
\caption{Positive and negative target regions ($N_{f}=10$).}
\vskip -0.5in
\label{fig:pos_neg_regions}
\end{center}
\end{wrapfigure}
We use CLIPseg \cite{lueddecke22_cvpr} to extract the target regions for localized editing. Specifically, CLIPSeg takes an image of size $352 \times 352$ as input, so we resize the $S \times S$ size rendered original patch $I^o(\theta, \mathbf{p})$ to $352 \times 352$ before feeding it to CLIPSeg. We then estimate the target region $M$ using the source text and the base text $\textit{`photo'}$. The dilatation operations with a kernel of size $5 \times 5$ are applied to obtain positive ($M_{+}$) and negative target ($M_{-}$) regions with $N_{f}$ and $2N_{f}$ times as shown in Figure \ref{fig:pos_neg_regions}. After this step, the regions are resized to fit the rendered image size $S \times S$ and the rest of the process is performed. We use the consistent notation $M_{+}$ and $M_{-}$ for readability. 

\vskip 0.3in
\subsection{Hyperparameter for Region Loss}
We observed that if adding densities is included in the editing operations, Blending-NeRF has difficulty making densities at initial when giving the same weights (\textit{i.e}. $\lambda_{+}=1$) to the positive and negative regions in the loss $\mathcal{L}_{\text{region}}$ of Eq.~(\ref{eq:loss_region}), especially for the small target region. To compensate for this, we set $\lambda_{+}$ by the ratio $r$ as:

\vspace{-10px}
\begin{equation}
\label{eq:ratio}
\begin{split}
r= \max(30, (S^2 - \textit{area of } M_{+}) / (1+\textit{area of } M_{+}))
\end{split}
\end{equation}
Note that this method is only used for object editing task that involves the adding density operation.

\vskip 0.3in
\subsection{Patch Sampling for Training}
Given the sampled camera pose $\mathbf{p}$, the rays are sampled at even intervals to make an image patch covering the entire extent of the image plane.
Specifically, let the width and height of the image plane and the patch size be $W$, $H$, and $S$, respectively; the starting points along each axis are uniformly sampled by the following distributions: 
\begin{equation}
\label{eq:distribution}
\begin{split}
\mathcal{U}(0, \lfloor W/S \rfloor + (W\mod{S})-1) \\
\mathcal{U}(0, \lfloor H/S \rfloor + (H\mod{S})-1). \\
\end{split}
\end{equation}
From these starting points, image patches are rendered at even intervals with horizontal stride $\lfloor W/S \rfloor$ and vertical stride $\lfloor H/S \rfloor$.
Finally, we can obtain $S \times S$ image and opacity patches.

\renewcommand{\thefootnote}{\arabic{footnote}}

\vskip 0.3in
\subsection{Training Time}
We train our model with 3k iterations on tasks that only change color or remove densities. In the tasks of adding densities with color change and the tasks of adding only densities, we iteratively train our model for 4k and 5k, respectively. Our method takes about 12 minutes to train 1k iterations on a single NVIDIA RTX A5000. It takes slightly longer than CLIP-NeRF~\cite{wang2022clip}\footnote{\url{https://github.com/cassiePython/CLIPNeRF}} which takes about 9 minutes to train 1k iterations due to our blending operations and additional objectives. Note that our approach can be extended to other more efficient 3D representation methods such as Instant-NGP~\cite{muller2022instant}. The detailed experiments incorporated with Instant-NGP are described in Section \ref{section:ngp_exp}.

\newpage
\begin{wrapfigure}{r}{0.35\textwidth}
\begin{center}
\vskip -0.7in
\centerline{\includegraphics[width=0.35\columnwidth]{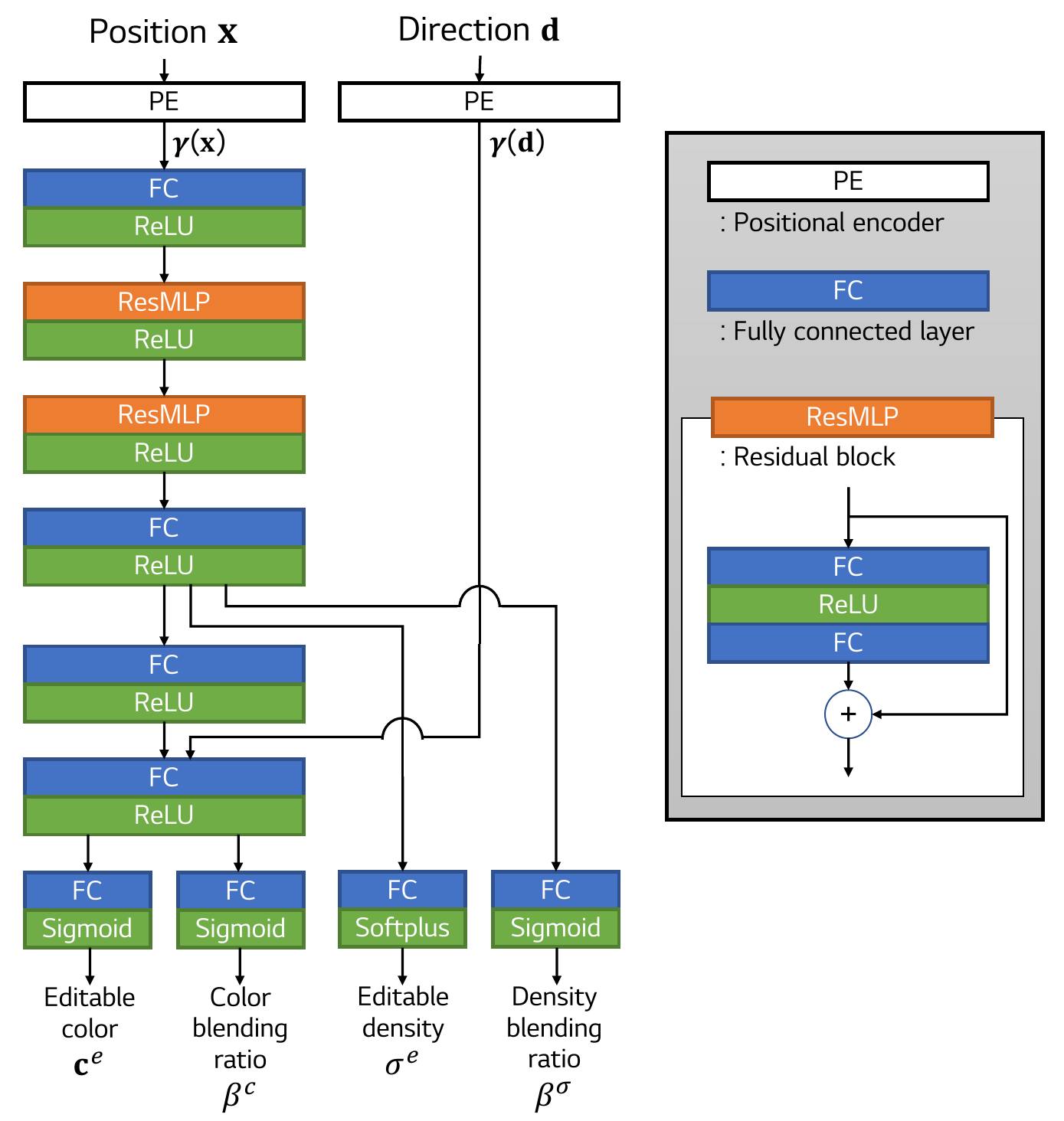}}
\caption{Editable NeRF Architecture.}
\vskip 0.5in
\label{fig:editable_nerf}
\end{center}
\vskip -1.5in
\end{wrapfigure}

\subsection{Annealing Thresholds}
For the editable amount constraint, we anneal two thresholds: $\tau^{add}$ and $\tau^{change}$. We use different target threshold values for each object editing task. Generally, $\tau^{add}$ is linearly annealed from 0.8 to the target value during the first 100 steps and remains the target value for the rest of the steps. Similarly, $\tau^{change}$ is annealed from 0.5--0.15 to the target value.

\vskip 0.3in
\subsection{Editable NeRF Architecture}
The detailed architecture of editable NeRF using residual blocks is shown in Figure \ref{fig:editable_nerf}. The editable NeRF extends NeRF to produce a density $\sigma^e \in [0, \infty)$ and color $\mathbf{c}^e \in [0, 1]^3$, in addition to two blending ratios $\beta^c \in [0, 1]$ and $\beta^\sigma \in [0, 1]$, respectively.

\vskip 0.6in
\section{Ablation Study}
\begin{figure}[h!]
\begin{center}
\centerline{\includegraphics[width=0.833\columnwidth]{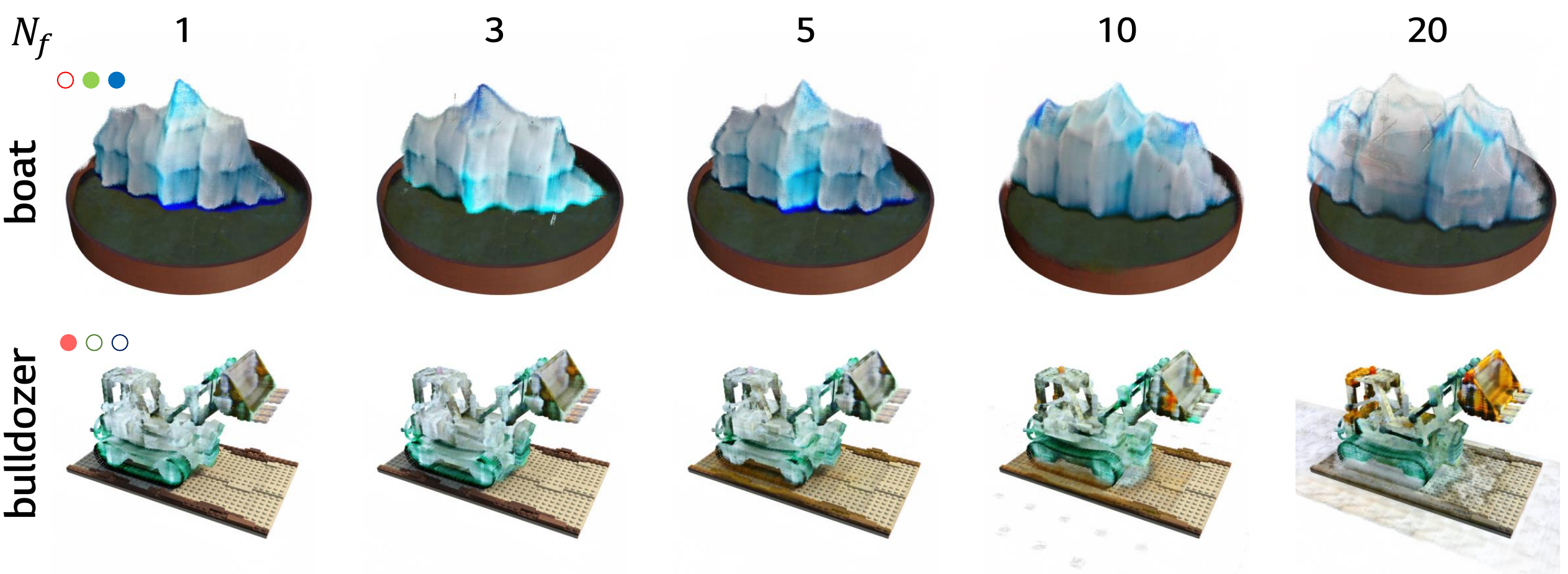}}
\caption{Ablation study on $N_{f}$. The $\textit{`boat'}$ object is edited to $\textit{`iceberg'}$ (upper row) with $\tau^{\text{add}}=0.35$ and $\tau^{\text{remove}}=0.05$, and the $\textit{`bulldozer'}$ is edited to $\textit{`marble bulldozer'}$ with $\tau^{\text{change}}=0.2$ (bottom row). 
}
\vskip -0.4in
\label{fig:ablation_N_filter}
\end{center}
\end{figure}

\paragraph{Dilatation operations}
To investigate the effect of the number of dilatation operations used in localizing the target, we qualitatively compare the editing results with different $N_{f}$ as shown in Figure \ref{fig:ablation_N_filter}. As shown in the upper row, the larger the number of dilatation operations, the larger the object is created as the target region grows. In the experiment that only changes the color, if $N_{f}$ becomes too large, noise appears on the object as shown in the bottom row.

\begin{figure}[h!]
\begin{center}
\centerline{\includegraphics[width=1.0\columnwidth]{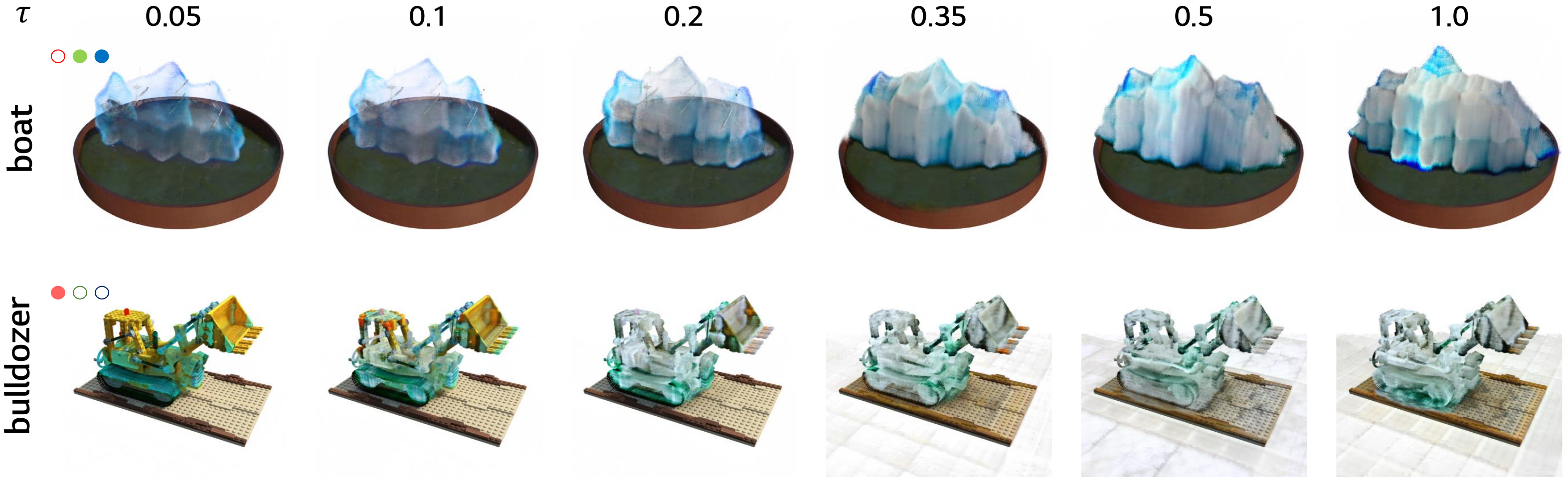}}
\caption{Ablation study on $\tau$. The $\textit{`boat'}$ object is edited to $\textit{`iceberg'}$ (upper row) with $N_{f}=10$ and $\tau^{\text{remove}}=0.05$, and the $\textit{`bulldozer'}$ is edited to $\textit{`marble bulldozer'}$ with $N_{f}=3$ (bottom row).}
\vskip -0.4in
\label{fig:ablation_tau}
\end{center}
\end{figure}

\vskip -0.2in
\paragraph{Constraining the amount of editing}
We also analyze the effect of the thresholds $\tau^{add}$ and $\tau^{change}$ used in constraining the amount of editing. As shown in the upper row of Figure \ref{fig:ablation_tau}, the larger the threshold $\tau^{add}$ for adding densities, the denser the object is created. Similarly, in an experiment that only changes color, the amount of change in the object is limited by the threshold $\tau^{change}$.

\paragraph{Image and text augmentations}
We augment text and images to improve the editing quality. Specifically, we augment original, editable, and blended images using differential \cite{zhao2020diffaugment} and random perspective augmentations in the same order as in \cite{lee2022understanding}. Differential augmentations include color jittering, translation, and cut-out with the same setting as \cite{zhao2020diffaugment}. In the case of random perspective augmentation, we set the distortion scale as 0.4 with a probability of 0.5. From a rendered image, we obtain 24 augmented images, including the rendered image itself, and feed them to the image encoder of CLIP. We use the same templates for text augmentations as \cite{bar2022text2live} before feeding the source and target text to the text encoder of CLIP. A random number of randomly selected text templates are applied to the source and target texts to form the augmented texts.

\begin{figure}[h!]
\begin{center}
\centerline{\includegraphics[width=0.833\columnwidth]{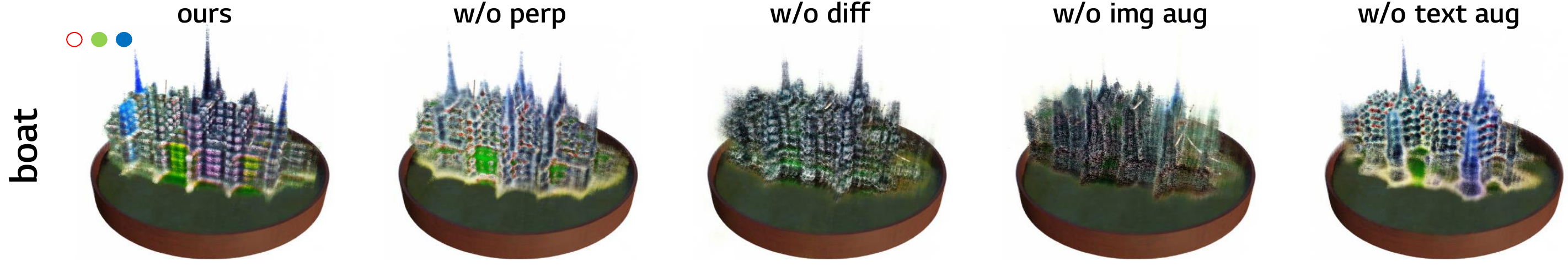}}
\caption{Ablation study on augmentations. The $\textit{`boat'}$ object is edited to $\textit{`fantasy modern city'}$.}
\vskip -0.35in
\label{fig:ablation_augmentation}
\end{center}
\end{figure}

We analyze the effect of augmentations by comparing the results when each augmentation is removed from the overall methods. As shown in Figure \ref{fig:ablation_augmentation}, without augmenting the rendered images (w/o img aug), the added object has quality degradation, such as blurred and ambiguous boundaries. We also noticed that among differential and perspective augmentations, the former has more effect on the quality improvements (w/o diff). Similarly, without text augmentations (w/o text aug), there is a slight deterioration in quality.

\vskip 0.3in
\section{User Study}
We conducted a user survey to evaluate the performance of our approach against the other three baselines. We asked 30 users to evaluate 20 randomly selected pairs from target text and rendered image pairs in a total of 60 scenes. The rendered images consist of four edited images rendered using each model, including all baseline results and ours, in addition to the original image. For a fair comparison, we randomly shuffled the order of the edited images. Then, each user was asked to assign a score (1--10) for each edited image based on the following criterion: ``How well is the image edited to match the target text relative to the amount the original object has changed?". We report the mean scores in Table~\ref{tab:human_evaluation}. Our approach obtained the highest user score, demonstrating once again the super performance of Blending-NeRF for text-driven editing.

\vspace{15px}
\begin{table}[h!]
    \begin{center}
    \begin{small}
    {
    \footnotesize
    \begin{adjustbox}{max width=0.99\columnwidth}
        \begin{tabular}{c c c c c}
        \toprule
          & CLIP-NeRF-\textit{c} & CLIP-NeRF-\textit{f} & CLIP-NeRF-\textit{D} & Ours\\
        \midrule
          score & $3.04$ & $3.99$ & $4.80$ & $\textbf{7.17}$ \\
        \bottomrule
        \end{tabular}
    \end{adjustbox}
    }
    \end{small}
    \end{center}
\caption{
Human assessment result for comparison with baselines. We report the mean score indicating the precision of text-driven editing by users. Our method outperforms all baselines.}
\label{tab:human_evaluation}
\end{table}

\newpage

\section{Experiments Using User-Provided Mask}
\label{user_provied_mask}

\begin{wrapfigure}{r}{0.41\textwidth}
\begin{center}
\vskip -0.3in
\centerline{\includegraphics[width=0.41\columnwidth]{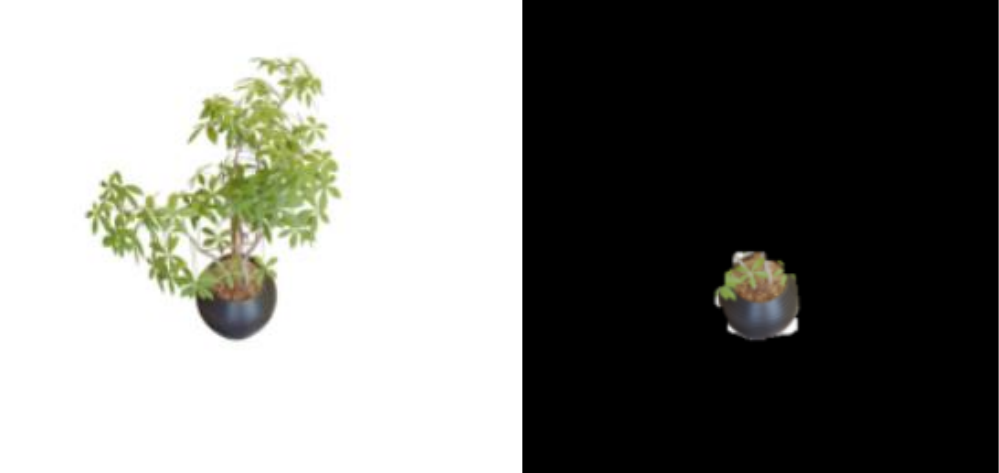}}
\caption{Inaccurate segmentation result for an image and a queried text $\textit{`brown-jar'}$. This incorrect segmentation may reduce the quality of localized object editing.}
\vskip -0.2in
\label{fig:segmentation_error}
\end{center}
\end{wrapfigure}

\begin{figure}[h!]
\vskip 0.2in
\begin{center}
\centerline{\includegraphics[width=1.0\columnwidth]{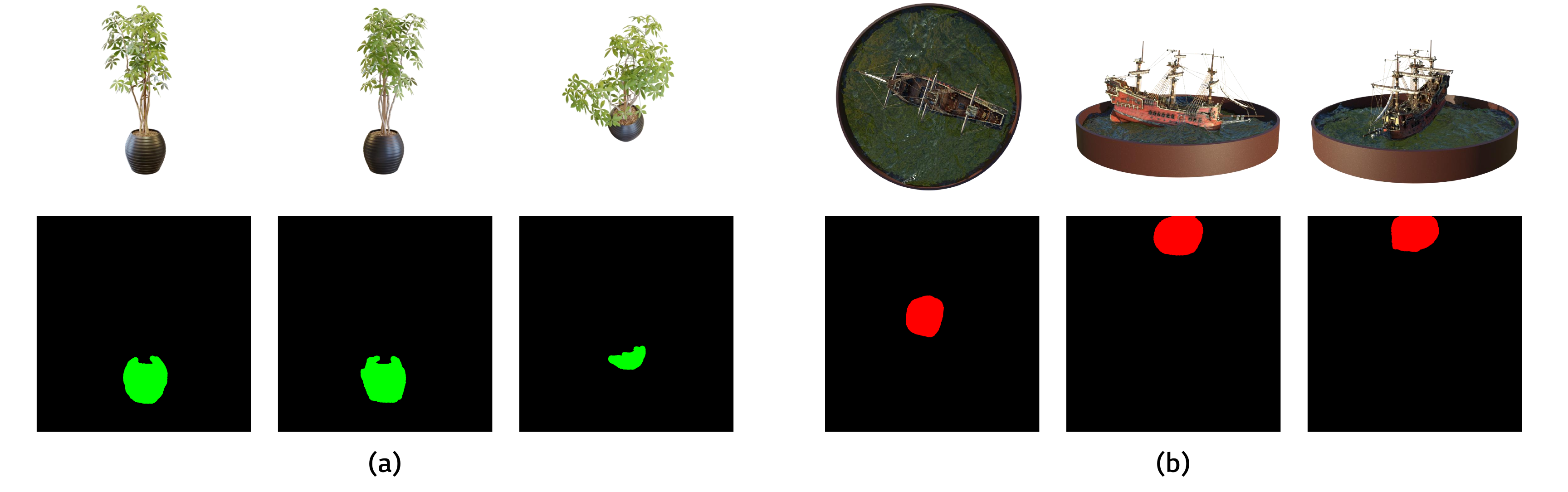}}
\vskip -0.1in
\caption{Users can utilize the user-provided image masks (a) to remove noise, or (b) to edit a specific region on the 3D objects.}
\vskip -0.2in
\label{fig:masking}
\end{center}
\end{figure}

\begin{figure}[h!]
\begin{center}
\centerline{\includegraphics[width=0.7\columnwidth]{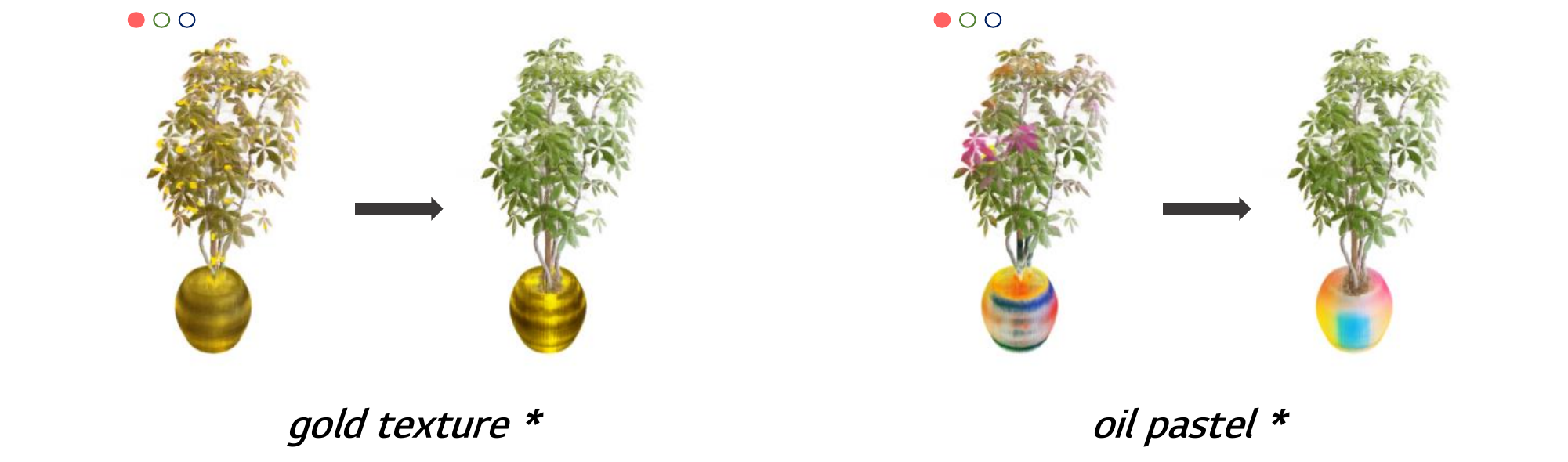}}
\caption{Examples of removing noise. By using user-provided masks, existing noise can be removed.}
\vskip -0.2in
\label{fig:masking_noise}
\end{center}
\end{figure}

\begin{figure}[h!]
\begin{center}
\centerline{\includegraphics[width=1\columnwidth]{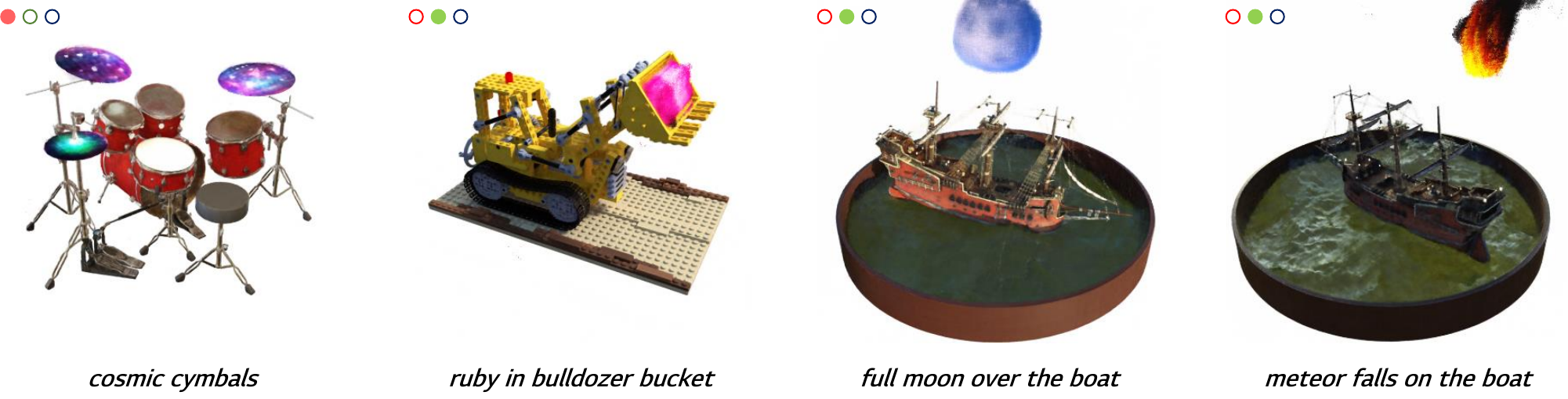}}
\caption{Examples of object editing using user-provided masks. If it is difficult to specify the target region with the source text, a user-provided mask can be used.}
\vskip -0.1in
\label{fig:masking_add}
\end{center}
\end{figure}

Additionally, we introduce a method using a user-provided target region for localized editing. We use this method to address two issues related to the target region. The first is that the text-based image segmentation is not accurate as shown in Figure \ref{fig:segmentation_error}. The second issue is that users may not be able to specify the target region with a text prompt. To address these issues, we manually set the target region mask to be edited from three appropriate viewpoints as shown in Figure \ref{fig:masking}. Then, in the middle of training (i.e. once every 5 training iterations), we use the user-provided masks.
Specifically, when the user-provided masks are given to remove noise as shown in \ref{fig:masking}-(a), the masks are used once out of 5 training iterations, and the existing segmentation results are used for the rest. On the other hand, if masks are provided to edit a specific region that cannot be specified by a text (\eg, \textit{`\textbf{over} the boat'}) as shown in \ref{fig:masking}-(b), image segmentation is not used. We present some editing results using two kinds of user-provided masks: masks for removing noise and masks for editing a specific area. First, as shown in Figure \ref{fig:masking_noise}, by giving accurate masks for editing, the existing noise caused by inaccurate segmentation can be removed. Second, as shown in Figure \ref{fig:masking_add}, by only using masks specifying the target regions, Blending-NeRF can accurately change the colors of the object (\eg, $\textit{`cymbals'}$ to $\textit{`cosmic cymbals'}$) and add densities to the scene (\eg, $\textit{`bulldozer'}$ to $\textit{`ruby in bulldozer bucket'}$).

\newpage

\section{Applying Blending-NeRF to Instant-NGP}
\label{section:ngp_exp}
Our novel layered architecture and blending operations for 3D object editing can be applied to other neural scene representations such as Instant-NGP. To demonstrate our approach can be incorporated with other 3D representation methods, we further experimented using Instant-NGP as a backbone which utilizes multi-resolution hash encoding to represent high-frequency details of a scene with low computational cost.

\begin{figure}[h!]
\vskip -0.15in
\begin{center}
\centerline{\includegraphics[width=0.92\columnwidth]{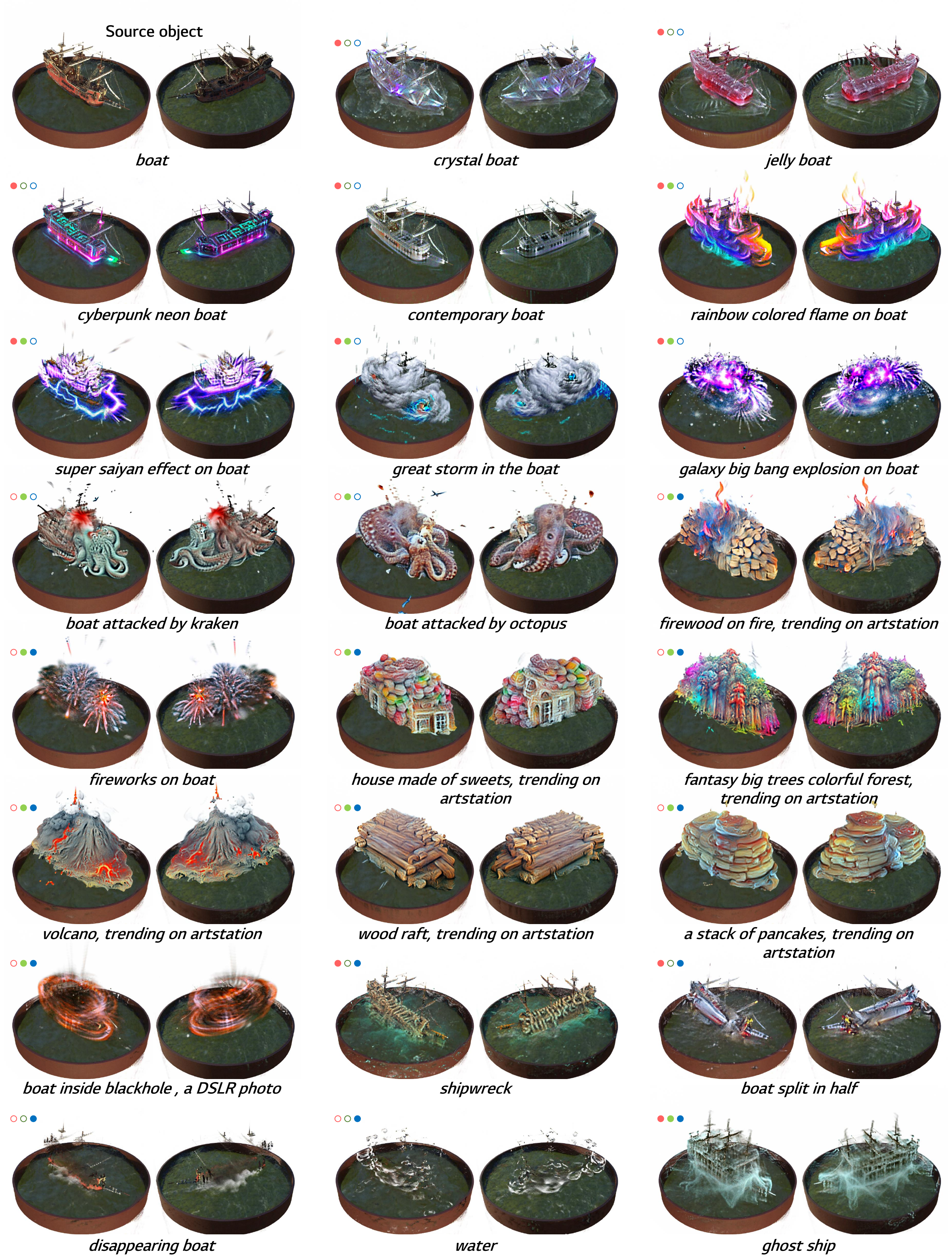}}
\caption{Editing results of our method integrated with Instant-NGP~\cite{muller2022instant}. The source object (\textit{`boat'}) at the top left is edited into each object using the target text at the bottom of each scene. For example, the result in the last column of the second row is edited from \textit{`boat'} to \textit{`rainbow colored flame on boat'}, combining color change ({\color{RubineRed} $\bullet$}) and density addition ({\color{LimeGreen} $\bullet$}).}
\centerline{{\color{RubineRed} $\bullet$} : changing colors \qquad {\color{LimeGreen} $\bullet$} : adding densities \qquad {\color{NavyBlue} $\bullet$} : removing densities.}
\label{fig:ngp_boat}
\end{center}
\vskip -2in
\end{figure}

We used the PyTorch and CUDA based reimplementation\footnote{\url{https://github.com/kwea123/ngp_pl}} of Instant-NGP, and the default settings in the codes were used. We applied the regularization loss $\mathcal{L}_{\text{reg}}$ at the start of training and set its weight as $\lambda_{\text{3}}=1.0$. The initial learning rate was set to $1\times10^{-3}$, and the rest of the hyperparameters were used the same. We trained all the edited scenes for 2k iterations with the patch size $S=128$, taking about 9 minutes to edit each scene on a single NVIDIA RTX A5000. Figure \ref{fig:ngp_boat} shows the locally edited 3D objects on the \textit{ship} scene with a wide range of target texts. The experimental results confirm that our approach can be incorporated with the other 3D representation methods.

\newpage

\end{document}